\documentclass[onecolumn]{robbyant}           %

\metadata[Website]{\url{https://technology.robbyant.com/lingbot-va}}
\metadata[Github]{\url{https://github.com/robbyant/lingbot-va}}
\metadata[Checkpoints]{\url{https://huggingface.co/robbyant/lingbot-va}}
\newcommand\methodname{LingBot-VA}

\newcommand{\method}{\texttt{\methodname}\xspace}

\usepackage{tikz}
\usetikzlibrary{tikzmark, calc}
\usepackage{makecell} 

\newcommand{\Func}[1]{\textsc{#1}}
\newcommand{\Var}[1]{\texttt{#1}}

\newcommand{\DrawScopeLine}[5]{
    \begin{tikzpicture}[overlay, remember picture]
        \coordinate (TopPoint) at ($ (pic cs:#1) + (-#3, #4) $);
        
        \coordinate (BottomRef) at ($ (pic cs:#2) + (0, -#5) $);
        
        \draw[gray!50, thick] (TopPoint) -- (TopPoint |- BottomRef);
    \end{tikzpicture}
}

\algblockdefx[MainLoop]{MainLoop}{EndMainLoop}{\textbf{loop}}{\textbf{end loop}}

\newcommand{\IndBranch}{\hspace{1.5em}} 
\newcommand{\IndCode}{\hspace{3.0em}}   
\newcommand{\IndInner}{\hspace{4.5em}}

\title{Causal World Modeling for Robot Control}

\author{
\begin{center}
    Lin Li$^{*}$ \quad
    Qihang Zhang$^{*\dagger}$ \quad
    Yiming Luo$^{*}$ \quad
    Shuai Yang \quad
    Ruilin Wang \quad
    Fei Han \quad
    \\[5pt]
    Mingrui Yu \quad
    Zelin Gao \quad
    Nan Xue \quad
    Xing Zhu \quad
    Yujun Shen \quad
    Yinghao Xu$^\ddagger$
    \\[12pt]
    $^{*}$Equal Contribution \qquad
    $^{\dagger}$Project Lead \qquad
    $^{\ddagger}$Corresponding Author
\end{center}
}

\begin{document}

\abstract{%
This work highlights that video world modeling, alongside vision-language pre-training, establishes a fresh and independent foundation for robot learning.
Intuitively, video world models provide the ability to ``imagine'' the near future by understanding the causality between actions and visual dynamics.
Inspired by this, we introduce \method, an autoregressive diffusion framework that learns frame prediction and policy execution simultaneously.
Our model features three carefully crafted designs:
(1) \textbf{\textit{a shared latent space}}, integrating vision and action tokens, driven by a Mixture-of-Transformers (MoT) architecture,
(2) \textbf{\textit{a closed-loop rollout mechanism}}, allowing for ongoing acquisition of environmental feedback with ground-truth observations,
(3) \textbf{\textit{an asynchronous inference pipeline}}, parallelizing action prediction and motor execution to support efficient control.
We evaluate our model on both simulation benchmarks and real-world scenarios, where it shows significant promise in long-horizon manipulation, data efficiency in post-training, and strong generalizability to novel configurations.
The code and model are made publicly available to facilitate the community.
}

\maketitle

\justifying
\section{Introduction}
\label{sec:introduction}

\begin{figure}[t]
    \centering              
    \includegraphics[width=\textwidth]{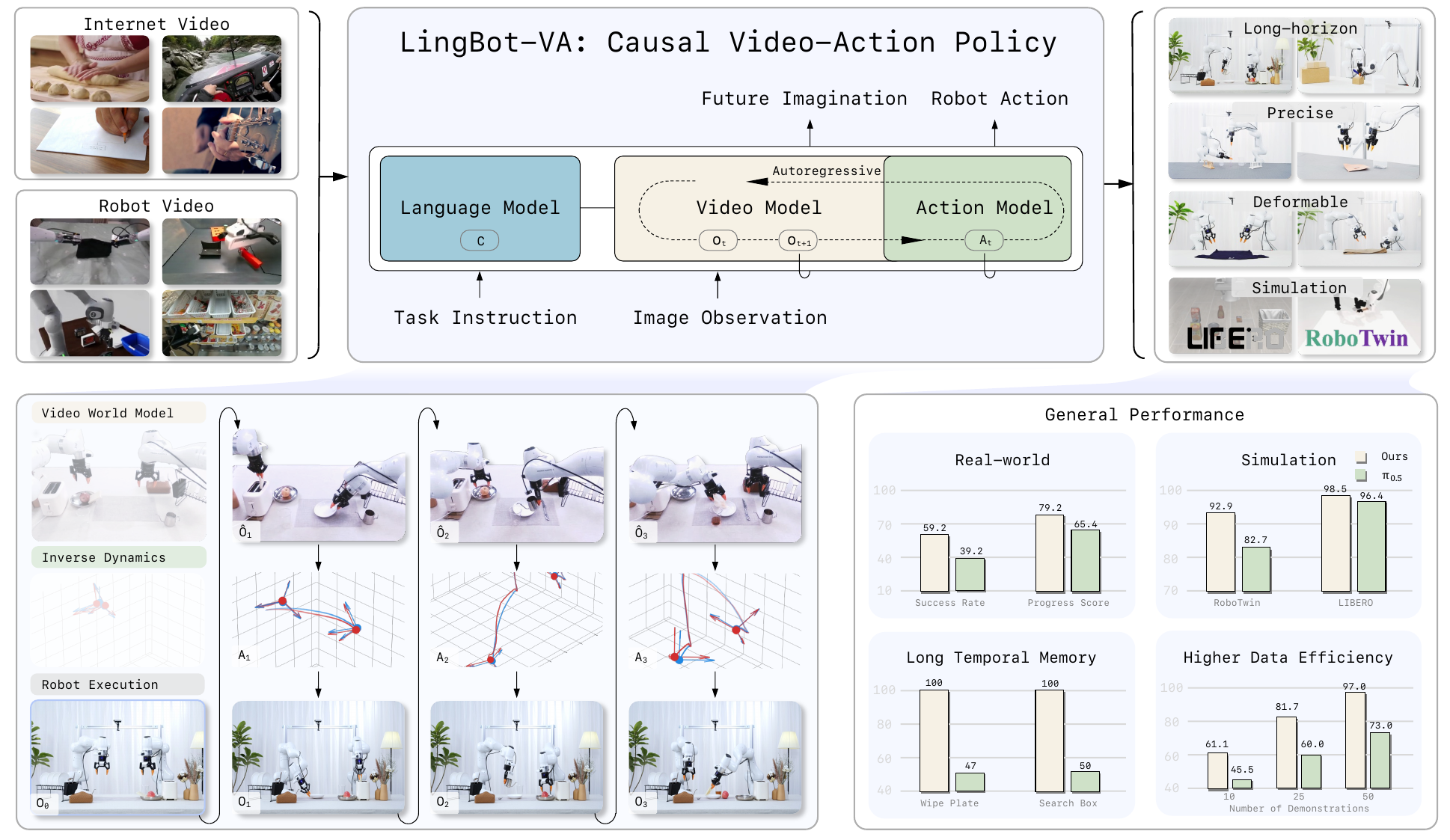}
    \caption{\method\textbf{: An Autoregressive World Model for Robotic Manipulation.}
    (1) \textbf{Pretraining:} \method is pretrained on diverse in-the-wild videos and robot action data, enabling strong generalization across scenes and objects.
    (2) \textbf{Comprehensive Evaluation:} We conduct extensive experiments on real-world tasks (long-horizon, deformable objects, and precision manipulation) and simulation benchmarks, significantly outperforming state-of-the-art methods including $\pi_{0.5}$.
    (3) \textbf{Versatile Capabilities:} Beyond policy learning, our model supports visual dynamics prediction and inverse dynamics inference from robot videos.
    (4) \textbf{Emergent Properties:} Our causal world modeling approach exhibits long-range temporal memory and strong few-shot adaptation ability.}
    \label{fig:teaser}
\end{figure}

Vision-Language-Action (VLA) models have emerged as a promising paradigm for general-purpose robotic manipulation~\cite{rt1,rt2,kim2024,pi0}, demonstrating impressive capabilities in grounding linguistic instructions into visual perceptions across diverse objects and unstructured environments.
However, beneath their apparent success lies a significant challenge: \emph{representation entanglement}.
Most existing VLAs adopt a feedforward paradigm that maps current observations to action sequences~\cite{diffusionpolicy,act}, requiring a single neural network to simultaneously learn visual scene understanding, physical dynamics, and motor control from a unified supervision signal.
This entanglement can create a bottleneck---the model must compress heterogeneous knowledge, ranging from high-dimensional visual semantics to low-dimensional motor commands, into a shared representation space.
This often leads to limited sample efficiency and suboptimal generalization.
Without explicit modeling of environmental evolution~\cite{dreamerv3,daydreamer,tdmpc2}, reactive policies may rely on pattern matching rather than a principled understanding of physical dynamics.

Recent attempts to bring world modeling into robotic policies span interactive neural simulators (e.g., UniSim~\cite{unisim}), chunk-based video-action diffusion models (e.g., UVA~\cite{uva} and UWM~\cite{uwm}), and offline video generators for subgoal synthesis (e.g. Gen2Act~\cite{gen2act}, Act2Goal~\cite{zhou2025act2goal}).
While conceptually appealing, these approaches face three primary limitations for effective closed-loop control.
First, \emph{the reactivity gap}: chunk/open-loop generation often rolls out long segments without incorporating real-time feedback, making it hard to adapt to disturbances.
Second, \emph{limited long-term memory}: chunk-wise generation can introduce inconsistencies over long horizons when history is not persistently cached.
Third, \emph{causality}: bidirectional attention within a segment allows future tokens to influence past predictions, which diverges from the causal nature of physical reality where the present depends only on the past.
These observations motivate an autoregressive formulation for robust closed-loop reasoning.

We propose \method, an \emph{autoregressive diffusion} world model that addresses these limitations through a unified video-action framework.
Unlike autoregressive language models that predict discrete tokens, our model operates in a continuous latent space via flow matching~\cite{flowmatching,rectifiedflow}, autoregressively generating chunks of video and action representations through iterative denoising.
While our approach conceptually separates visual dynamics prediction and action decoding~\cite{vidar,vpp}, the key architectural insight is to \emph{interleave} video and action tokens into a single autoregressive sequence.
Both modalities are jointly processed through a Mixture-of-Transformers (MoT) architecture~\cite{mot} with shared attention.
Within this unified autoregressive generation process, latent imagination and action inference occur jointly: at each autoregressive step, the model generates predicted future visual states through iterative denoising while simultaneously decoding the corresponding actions, allowing both streams to mutually condition on one another.
This integration, built upon a large-scale pretrained video diffusion backbone~\cite{wan2024video}, offers several advantages:
\emph{(i) Reactive AR loop}: because video and action tokens form a unified sequence, each autoregressive step allows the system to recalibrate based on the latest real-world observation, enabling timely adjustments to both the predicted future and motor commands;
\emph{(ii) Persistent context through KV-cache}: the cached key-value pairs preserve the interleaved video-action trajectory, providing a rich context that helps mitigate temporal drift;
\emph{(iii) Causal consistency}: causal attention masking over the unified sequence ensures that both predicted visual states and action commands are governed by preceding states, respecting the temporal arrow of physical dynamics.
By incorporating real-world observations at each step, this formulation helps mitigate the \emph{distribution drift} that often affects open-loop methods in long-horizon tasks.

A primary challenge in deploying large-scale autoregressive video-action models is inference latency; generating high-fidelity video tokens through iterative denoising is computationally intensive.
We address this through two complementary strategies.
First, we introduce \emph{Noisy History Augmentation}, a training scheme that enables \emph{partial denoising} at inference time.
The key insight is that action decoding does not always require pixel-perfect reconstruction; instead, it can rely on robust semantic structures.
By training the action decoder to predict from partially noisy latent representations, we significantly reduce the computational overhead while maintaining precise action prediction.
Second, we design an \emph{asynchronous coordination} pipeline that overlaps computation with execution: while the robot executes current actions, the world model predicts future visual states and plans subsequent sequences.
This parallelized architecture, combined with variable chunk-size training, facilitates high-frequency closed-loop control without compromising prediction quality.

We evaluate \method across diverse manipulation tasks in both simulation and real-world environments.
Our method demonstrates competitive performance compared to state-of-the-art VLA policies, particularly in long-horizon tasks requiring temporal consistency.
Our contributions are summarized as follows:

\begin{itemize}
    \item \textbf{Autoregressive Video-Action World Modeling:} We introduce an autoregressive diffusion framework that \emph{architecturally unifies} visual dynamics prediction and action inference within a single interleaved sequence while maintaining their \emph{conceptual distinction}. This formulation supports persistent memory through KV cache and causal consistency via attention masking.

    \item \textbf{Mixture-of-Transformers Architecture with Asynchronous Execution:} We design a dual-stream MoT architecture with asymmetric capacity and introduce a partial denoising strategy combined with asynchronous coordination to enable efficient robotic control.

     \item \textbf{Superior Long-Horizon and Precision Performance:}
    Extensive real-world and simulation experiments demonstrate consistent state-of-the-art performance, with particularly strong improvements on long-horizon and high-precision manipulation tasks. Our method also achieves significantly improved sample efficiency and strong generalization to novel scenes and object configurations. 

\end{itemize}

\section{Preliminary}

\subsection{Flow Matching}

Flow matching~\cite{flowmatching,rectifiedflow,cfm} is a continuous-time generative modeling framework that learns to transform a simple
source distribution (e.g., Gaussian noise) to a target data distribution through a continuous flow.
Given a data sample $x_1$ and a noise sample $\epsilon \sim \mathcal{N}(0, I)$, flow matching defines
a time-dependent vector field $v_s: \mathbb{R}^d \times [0,1] \to \mathbb{R}^d$ that describes the
instantaneous velocity of particles flowing from $\epsilon$ to $x_1$.
The trajectory $x^{(s)}$ evolves according to the ordinary differential equation (ODE):
\begin{equation}
\frac{dx^{(s)}}{ds} = v_s(x^{(s)}), \quad x^{(0)} = \epsilon \sim \mathcal{N}(0, I),
\end{equation}
where $s \in [0, 1]$ denotes the flow time.

The model is trained to predict this vector field by minimizing:
\begin{equation}
\mathcal{L}_{\text{FM}} = \mathbb{E}_{s, \epsilon, x_1} \left[ \| v_\theta(x^{(s)}, s) - \dot{x}^{(s)} \|^2 \right],
\end{equation}
where $\dot{x}^{(s)}$ is the true velocity along the interpolation path, typically defined as $x^{(s)} = (1-s)\epsilon + sx_1$,
giving $\dot{x}^{(s)} = x_1 - \epsilon$.

At inference, samples are generated by solving the learned ODE from $s=0$ to $s=1$:
\begin{equation}
x_1 = \epsilon + \int_0^1 v_\theta(x^{(s)}, s) \, ds.
\end{equation}

\subsection{Video Generation with Conditional Flow Matching}

Recent video generation models~\cite{sora,kling,veo,wan2024video} leverage flow matching to generate videos conditioned on text or images.
These models operate in the latent space of pretrained video autoencoders, where visual observations
are encoded as latent representations $z_t = E(o_t)$ using encoder $E$ (e.g., from video diffusion models).

Given a conditioning signal $c$ (text prompt or initial image), the flow matching model learns to
generate a sequence of latent video frames $\mathbf{z} = \{z_1, \ldots, z_T\}$ by predicting the vector field:
\begin{equation}
v_\theta(\mathbf{z}^{(s)}, s \mid c) = \frac{d}{ds} \mathbf{z}^{(s)},
\end{equation}
where $s \in [0, 1]$ is the flow time and $\mathbf{z}^{(s)}$ represents the latent video at flow step $s$.
The generation process starts from noise $\mathbf{z}^{(0)} = \epsilon \sim \mathcal{N}(0, I)$ and integrates the learned
vector field to obtain the final latent video $\mathbf{z}^{(1)}$, which is then decoded to pixel space.
This bidirectional generation framework enables flexible synthesis from text descriptions or seed images.

\begin{figure}[t]
    {\centering
    \includegraphics[width=\textwidth]{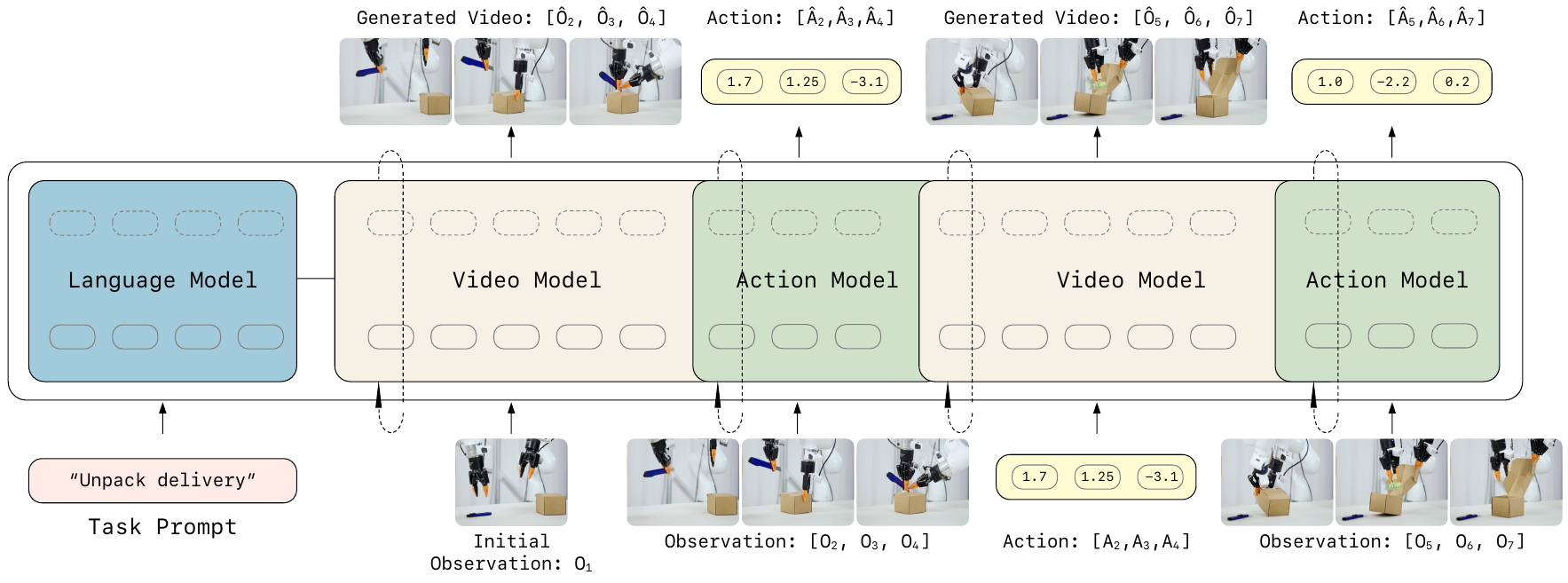}}

    \caption{\raggedright \textbf{Framework overview}: \method is conditioned by \textit{autoregressive diffusion} for unified \textit{video-action world modeling}. 
    We leverage a dual-stream \textit{Mixture-of-Transformers (MoT)} architecture that interleaves video and action tokens within a single sequence. 
    At each autoregressive step, the video stream (initialized from Wan2.2-5B) first predicts future latent visual states via \textit{flow matching}. 
    Then the action stream decodes corresponding actions through \textit{inverse dynamics} conditioning on the predicted visual transitions. 
    }
    \label{fig:overview}
\end{figure}

\section{Method}
\subsection{Problem Statement \& Approach Overview}

We study robotic manipulation as a sequential decision-making problem under partial observability.
At each timestep $t$, the agent receives a visual observation $o_t \in \mathcal{O}$ and executes
an action $a_t \in \mathcal{A}$, which induces a transition in the underlying physical world and
produces the next observation $o_{t+1}$.

\noindent\textbf{Vision-Language-Action (VLA) Policies.}
Most existing VLA policies learn a direct, reactive mapping from observation history to actions:
\begin{equation}
a_t \sim \pi_\theta(\cdot \mid o_t),
\end{equation}
through imitation learning on robot demonstration data.
While this end-to-end approach has shown impressive results, it suffers from a fundamental
coupling problem: the model must simultaneously learn visual scene understanding, physical dynamics,
and motor control from a single supervision signal of paired observations and actions.
This entanglement leads to poor sample efficiency and limited generalization, as the model struggles
to disentangle visual reasoning from action prediction without explicit dynamics modeling.

\noindent\textbf{Our Approach.}
Unlike VLA policies that directly learn action distributions, we adopt a world modeling perspective:
instead of learning $\pi(a_t \mid o_t)$, we predict how the visual world will evolve,
then infer actions based on these predictions.·
Our approach operates in two stages:
\begin{equation}
\begin{aligned}
&\text{(Stage 1) Visual dynamics prediction:} \quad &&o_{t+1} \sim p_\theta(\cdot \mid o_{\le t}), \\
&\text{(Stage 2) Inverse dynamics:} \quad &&a_t \sim g_\psi(\cdot \mid o_t, o_{t+1}).
\end{aligned}
\end{equation}
Stage 1 learns to predict future visual observations given observation history.
Stage 2 uses an inverse dynamics model to decode actions from desired visual transitions.
This decomposition enables Stage 1 to leverage large-scale video data for learning physical priors,
while Stage 2 only requires robot demonstrations to ground visual predictions in executable actions.

\noindent\textit{Method Overview.}
Figure~\ref{fig:overview} illustrates the details of our framework.
Our method consists of three key components, detailed in the following subsections:
\textit{(\S\ref{sec:world_model}) Autoregressive Video-Action World Modeling} describes how we
model visual dynamics in latent space and decode actions from predicted state transitions---this is the \emph{core formulation} of our approach;
\textit{(\S\ref{sec:architecture}) \methodname: Unified Architecture \& Training} presents our unified model
for video-action pretraining, including the architecture design and training objective---this is the \emph{instantiation} of our formulation;
\textit{(\S\ref{sec:inference}) Real-time Deployment \& Asynchronous Inference} introduces our deployment
strategy that enables real-time control through parallelized prediction and execution---this is the \emph{practical realization} for robotic control. 

\subsection{Autoregressive Video-Action World Modeling}
\label{sec:world_model}

Previous video world models either focus on open-ended video prediction~\cite{sora} or learn action-conditioned interactive environments~\cite{genie,genie2} primarily for game or simulation domains, which may not directly transfer to precise robotic manipulation.
To leverage rich visual dynamics priors from video data for robot manipulation, we propose a unified video-action world modeling framework that jointly models visual observations and robot actions within a single autoregressive process.
Unlike prior approaches that either decouple video prediction from action inference~\cite{moto,vpp} or rely on bidirectional diffusion within segments~\cite{uwm}, our method unifies video and action within a single \emph{causal autoregressive} framework, enabling persistent memory through KV cache and seamless integration of real-time observations.

\noindent\textbf{World Dynamics with Autoregressive Modeling.}
Recent world models for robotics often adopt bidirectional video generation approaches~\cite{unipi,dreamitate,gen2act,pwa} or learn interactive simulators~\cite{unisim}, which face fundamental limitations for closed-loop control.
Open-loop methods that generate entire long sequences in one shot incur prohibitive computational cost
and cannot incorporate real-time feedback for error correction.
Chunk-based diffusion methods that generate video segments sequentially~\cite{vidar,uwm} suffer from two critical issues:
(1) they lack persistent memory across chunks, as each chunk is generated independently without access to the full history, leading to temporal inconsistencies and drift over long horizons;
(2) the bidirectional attention within each chunk violates causality, preventing seamless integration with real-time observations during execution.

The physical world, however, is inherently causal and autoregressive: the present state depends only on the past, and we cannot observe the future before it occurs.
This fundamental property motivates our autoregressive world modeling approach, which offers three critical advantages over chunk-based diffusion for robotic control:
\textit{(1) Persistent Memory}: by explicitly conditioning on the complete observation history through causal attention and KV cache, the model maintains long-term context and temporal coherence across the entire trajectory, avoiding the ``amnesia'' problem of chunk-based methods;
\textit{(2) Causal Consistency}: the unidirectional dependency structure naturally aligns with closed-loop execution, where new observations can be seamlessly incorporated as they arrive;
\textit{(3) Efficiency}: chunk-wise prediction with parallel generation within each chunk balances computational efficiency with autoregressive flexibility, enabling high-frequency control with real-time error correction.

We formalize this as an autoregressive process: at each step, the world model predicts the next chunk of $K$ video frames using conditional flow matching:
\begin{equation}
o_{t+1:t+K} \sim p_\theta( \cdot \mid o_{\leq t}),
\end{equation}
where tokens within each chunk are generated in parallel via bidirectional attention, while maintaining causal structure across chunks.
This chunk-wise formulation balances generation efficiency with autoregressive flexibility for closed-loop correction.

\noindent\textbf{Video-Action State Encoding.}
Operating directly on pixel-level video observations is computationally prohibitive due to the high dimensionality and redundancy of raw visual data.
We leverage a causal video VAE~\cite{wan2024video} to compress visual observations into compact latent tokens $z_t = E(o_t \mid o_{<t}) \in \mathbb{R}^{N \times C}$, where $N$ is the number of spatial tokens after passing into video VAE, and $C$ is the channel number.
By conditioning on previous latent states, the encoder maintains temporal coherence while processing observations sequentially, naturally aligning with our autoregressive world modeling framework.
To align robot actions with visual tokens, we project action vectors to token embeddings $a_t \in \mathbb{R}^{D}$ via a lightweight MLP $\phi(\cdot)$ where $D$ is the dimension of the video token after patchfication, enabling unified interleaving of visual and action tokens as in prior approaches~\cite{vidar, motus}.

\noindent\textbf{Latent Video State Transition.}
While standard video generation models predict future frames based solely on visual history, robotic manipulation requires accounting for the embodiment's physical state and interaction with the environment.
During deployment, the robot's state evolves through continuous interaction: each action modifies the embodiment's configuration (e.g., gripper position, joint angles), which in turn influences how the scene evolves.

In many manipulation settings, actions encode absolute pose information (e.g., end-effector poses in world coordinates), so the action history $a_{<t}$ effectively captures the trajectory of the embodiment's configuration.
Conditioning on action history thus provides knowledge of how the robot has moved and interacted with objects, consistent with prior action-conditioned video/world models~\cite{unisim,vidar,uwm}.
We extend our autoregressive formulation to condition on both observation and action histories:
\begin{equation}
z_{t+1:t+K} \sim p_\theta( \cdot \mid z_{\leq t}, a_{<t}),
\label{eq:vdm}
\end{equation}
where $z_t$ is the latent visual state and $a_t$ is the action token.
This enables the world model to ground predictions in the embodiment's state, ensuring that predicted observations reflect the robot's physical interaction with the scene.

\noindent\textbf{Inverse Dynamics for Action Decoding.}
Once the world model predicts future visual states, we leverage these predictions to plan actions.
Rather than directly predicting actions from current observations, we employ an inverse dynamics model that infers actions by conditioning on desired future observations, enabling the policy to reason about \textit{what action leads to a desired visual outcome}.

However, simply conditioning on the current and next states $(z_t, z_{t+1})$ is insufficient for accurate action prediction.
The action history $a_{<t}$ encodes the embodiment's state trajectory for determining feasible actions, while the observation history $z_{<t}$ provides temporal context for multi-step interactions (e.g., whether an object was previously grasped).
We therefore formulate inverse dynamics as:
\begin{equation}
a_{t:t+K-1} \sim g_\psi(\cdot \mid \hat{z}_{t+1:t+K}, z_{\leq t}, a_{<t}),
\label{eq:idm}
\end{equation}
where the inverse dynamics model $g_\psi$ takes as input the predicted chunk of visual states $\hat{z}_{t+1:t+K}$ inferred by Eq.~\ref{eq:vdm}, observation history $z_{\leq t}$, and action history $a_{<t}$.
This mirrors recent IDM-based policies~\cite{unipi,onexwm,mimicvideo,vidar,tian2025} that leverage future targets to infer feasible actions while maintaining consistency with embodiment dynamics.

\subsection{\methodname: Unified Architecture \& Training}
\label{sec:architecture}

\noindent\textbf{Architecture.}
To jointly model video and action generation, we leverage a dual-stream diffusion transformer architecture that performs conditional flow matching for autoregressive prediction.
Our model consists of two parallel transformer backbones: a video stream initialized from Wan2.2-5B (a large-scale pretrained video generation model with dimension $d_v$~\cite{wan2024video}), and an action stream with same depth but significantly smaller width $d_a \ll d_v$.
This asymmetric design is motivated by the observation that action distributions are inherently simpler than visual data requiring fewer parameters to model effectively while maintaining expressive capacity for visual dynamics.

\noindent\textit{Video Sparsification.}
Video frames exhibit significant temporal redundancy, especially in robotic manipulation where scenes evolve gradually.
We sparsify the video sequence by temporally downsampling frames by a factor of $\tau=4$, reducing visual tokens while improving efficiency~\cite{motus}.
Since actions evolve at higher frequency than visual changes, we interleave the downsampled video tokens with action tokens in temporal order: for each video frame $o_t$, we associate $\tau$ consecutive actions $\{a_{t,1}, a_{t,2}, \ldots, a_{t,\tau}\}$, forming a unified sequence $[z_t, a_{t,1}, a_{t,2}, \ldots, a_{t,\tau}, z_{t+1}, \ldots]$ for joint modeling.
This design means that predicting $K$ video frames corresponds to generating $\tau K$ actions, enabling high-frequency control while maintaining efficient video generation.

\noindent\textit{Mixture-of-Transformer Block.} To enable interaction while preserving modality-specific feature spaces, we employ a Mixture-of-Transformers (MOT) architecture~\cite{motus,mot,bagel}, where video and action tokens are processed by separate transformer blocks at each layer, then fused via cross-modal attention~\cite{motus}.
At each layer, the video and action streams independently compute their query, key, and value matrices using separate QKV projection matrices, maintaining distinct feature spaces for each modality.
To align dimensions for cross-modal fusion, action tokens are first projected to the video dimension via a linear layer, participate in joint self-attention, then projected back to their original dimension via a residual connection that preserves the action-specific representations.
This MOT design allows video and action to mutually influence each other through attention while maintaining separate parameterizations, preventing interference between modality-specific feature representations.
For action decoding, the final action stream outputs are mapped to low-dimensional action vectors via a linear projection head.

\noindent\textit{Action Network Initialization.}
Proper initialization of the action stream is critical for training stability and convergence.
We find that training the action network from scratch leads to unstable optimization and slow convergence, as the action tokens' output distribution initially diverges significantly from the video distribution, disrupting the joint attention mechanism.
To address this, we initialize the action network weights by interpolating the pretrained video weights according to the action dimension, then apply a scaling factor $\alpha = \sqrt{d_v / d_a}$ to preserve output variance, where $d_v$ and $d_a$ are the video and action dimensions.
This initialization strategy ensures that action tokens start with output distributions comparable to video tokens, stabilizing early-stage training and accelerating convergence.

\noindent\textit{Variable Chunk Size Training.}
To enable flexible deployment, we randomly sample the chunk size $K$ from a predefined range during training.
By training with variable chunk sizes (e.g., $K \in [1, 8]$), the model learns to generate coherent predictions across different temporal horizons.
At inference time, this allows freely selecting the chunk size to balance computational efficiency and planning horizon---larger chunks reduce the number of autoregressive steps but require longer per-step computation, while smaller chunks enable more frequent closed-loop correction.
In our experiments, we use $K=4$ for deployment as a practical trade-off.

\noindent\textbf{Teacher Forcing for Unified Video-Action Training.}
In \S\ref{sec:world_model}, we formulated both visual dynamics prediction (Eq. 7) and inverse dynamics (Eq. 8) as autoregressive modeling problems, where each prediction conditions on the history of observations and actions.
This unified autoregressive formulation enables a natural training strategy: we can treat the interleaved video-action sequence as a single unified sequence and train the model using standard next-token prediction, analogous to language modeling in NLP~\cite{transformer}.

\begin{wrapfigure}{r}{0.45\textwidth}
    \centering
    \vspace{-10pt}
    \includegraphics[width=0.45\textwidth]{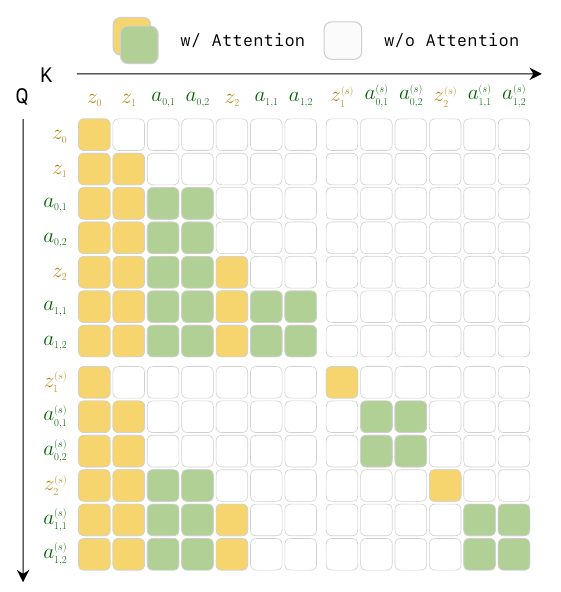}
    \caption{ \raggedright\textbf{Teacher Forcing Attention Mask}: Causal attention mask for unified video-action pretraining. Each token can only attend to preceding tokens in the temporal sequence.}
    \label{fig:attention_mask}
    \vspace{-10pt}
\end{wrapfigure}

Specifically, given an episode with interleaved tokens, we train the model to predict each token conditioned on all preceding tokens in the sequence.
This is implemented via teacher forcing: during training, we use ground-truth tokens from the dataset as context for predicting subsequent tokens, rather than model-generated predictions.
The causal dependency structure is enforced through attention masking (Figure~\ref{fig:attention_mask})---each token can only attend to tokens that appear earlier in the temporal sequence.

Importantly, teacher forcing is particularly well-suited for robotic manipulation: unlike pure generative modeling where it leads to train-test distribution mismatch, robot policies naturally retrieve real-world observations during deployment, directly matching the training regime.
This formulation offers two key benefits: (1) unifying video and action prediction under a single training objective enables end-to-end learning of world dynamics and action inference; (2) by processing episodes in parallel with causal attention masking, we efficiently optimize both components across all timesteps in a single forward pass.

\noindent\textit{Noisy History Augmentation.}
The primary bottleneck during inference remains video token generation---the number of video tokens are much larger than action tokens, and each requires multiple denoising steps through the flow matching process.
To address this, we introduce a noise augmentation strategy during training that enables \emph{partial denoising} at test time.
The key insight is that action prediction does not require fully denoised video representations; the inverse dynamics model can learn to extract action-relevant information from partially noisy video states.
Specifically, during training, we randomly augment the video history $z_{\leq t}$ with noise following the same interpolation scheme as flow matching:
\begin{equation}
\tilde{z}_{\leq t} = \begin{cases}
(1 - s_{\text{aug}}) \epsilon + s_{\text{aug}} z_{\leq t}, & p = 0.5, \quad s_{\text{aug}} \in [0.5, 1], \; \epsilon \sim \mathcal{N}(0, I) \\
z_{\leq t}, & 1 - p = 0.5
\end{cases}
\end{equation}
This augmentation trains the action decoder to predict actions from partially noisy video representations.

At inference time, this enables a significant speedup: instead of fully denoising video tokens from $s=0$ to $s=1$, we only need to denoise to $s=0.5$, halving the number of denoising steps for video generation while maintaining action prediction quality.

\noindent\textbf{Training Objective.}
We jointly optimize both video and action using flow matching with the noisy history augmentation described above.
For video tokens $z_t$, the dynamics loss supervises velocity field prediction conditioned on (potentially noisy) history:
\begin{equation}
\mathcal{L}_{\mathrm{dyn}} = \mathbb{E}_{t, s, z_{t+1}, \epsilon} \left[ \| v_\theta(z_{t+1}^{(s)}, s ,  \tilde{z}_{\leq t}, a_{<t} | c) - \dot{z}_{t+1}^{(s)} \|^2 \right],
\end{equation}
where $s \in [0,1]$ is flow time, $z_{t+1}^{(s)} = (1-s)\epsilon + sz_{t+1}$ with $\epsilon \sim \mathcal{N}(0, I)$, $\dot{z}_{t+1}^{(s)} = z_{t+1} - \epsilon$, $\tilde{z}_{\leq t}$ is the augmented history (Eq.~10), and $c$ is the language instruction.
For action tokens $a_t$, the inverse dynamics loss conditions on current and next observations:
\begin{equation}
\mathcal{L}_{\mathrm{inv}} = \mathbb{E}_{t, s, a_t, \epsilon} \left[ \| v_\psi(a_t^{(s)}, s, \tilde{z}_{\leq t+1}, a_{<t} | c) - \dot{a}_t^{(s)} \|^2 \right],
\end{equation}
where $a_t^{(s)} = (1-s)\epsilon + sa_t$ with $\epsilon \sim \mathcal{N}(0, I)$, $\tilde{z}_{t}, \tilde{z}_{t+1}$ are the (potentially noisy) current and next video tokens, and $c$ is the language instruction.
The complete objective is $\mathcal{L} = \mathcal{L}_{\mathrm{dyn}} + \lambda \mathcal{L}_{\mathrm{inv}}$.

\subsection{Real-time Deployment \& Asynchronous Inference}
\label{sec:inference}

\noindent\textbf{KV Cache for Efficient Autoregressive Inference.}
Our autoregressive formulation naturally enables KV cache acceleration during inference.
Since each prediction step conditions on the history of observations and actions, we cache the key-value pairs from previous tokens to avoid redundant computation.
At each autoregressive step, only the new tokens (current observation and predicted actions) require full attention computation, while cached history tokens are reused.
Algorithm~\ref{alg:kv_inference} describes the complete inference procedure with KV cache.

\begin{algorithm}[t]
\caption{KV Cache Inference}
\label{alg:kv_inference}
\begin{algorithmic}[1]
\Require Initial observation $o_0$, chunk size $K$, KV cache $\mathcal{C}$
\State $z_0 \gets E(o_0)$, $\mathcal{C} \gets \{z_0\}$
\State $t \gets 0$

\MainLoop
    \State Sample $\epsilon \sim \mathcal{N}(0, I)$ \Comment{Generate video chunk (integrate to $s=0.5$)}
    \State $\tilde{z}_{t+1:t+K} \gets \epsilon + \int_0^{0.5} v_\theta(z^{(s)}_{t+1:t+K}, s \mid \mathcal{C}) \, ds$
   
    \State Sample $\epsilon \sim \mathcal{N}(0, I)$ \Comment{Generate action chunk (integrate to $s=1$)}
    \State $a_{t:t+K-1} \gets \epsilon + \int_0^{1} v_\psi(a^{(s)}_{t:t+K-1}, s \mid \tilde{z}_{t:t+K}, \mathcal{C}) \, ds$
    
    \For{$i = t$ \textbf{to} $t+K-1$}
        \State Execute $a_i$, receive $o_{i+1}$ \Comment{Execute and collect observations}
        \State $z_{i+1} \gets E(o_{i+1})$
    \EndFor
    
    \State $\mathcal{C} \gets \mathcal{C} \cup \{z_{t+1:t+K}, a_{t:t+K-1}\}$ \Comment{Update KV cache}
    \State $t \gets t + K$
\EndMainLoop %
\end{algorithmic}
\end{algorithm}

\begin{figure}[t]
    {\centering
    \includegraphics[width=\textwidth]{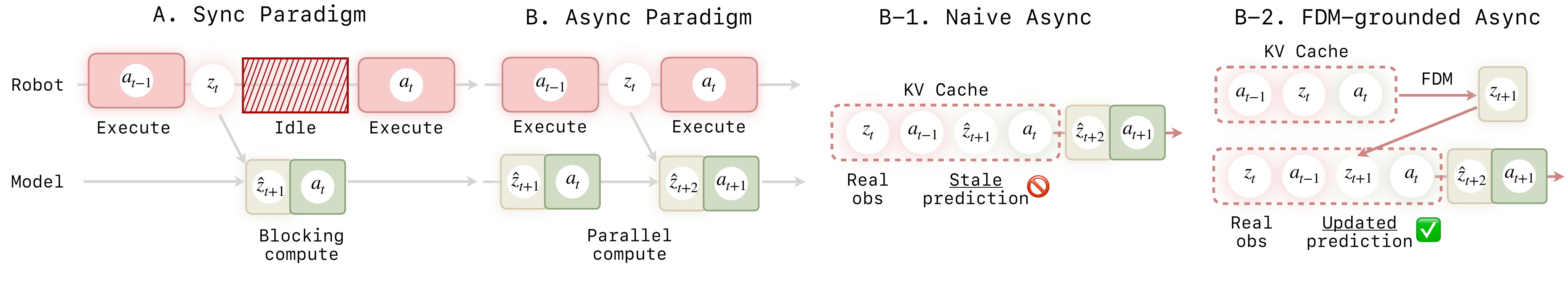}}

    \caption{\raggedright \textbf{Asynchronous pipeline design overview}: The traditional synchronous pipeline (A) suffers from delays caused by blocked computations, while the asynchronous pipeline (B) addresses this issue by enabling parallel computation and execution. However, a naive asynchronous implementation (B-1) relies on outdated visual predictions. In contrast, we improve and refine asynchronous prediction through forward dynamic prediction (B-2), which updates stale predictions with recent real-world observations.
    }
    \label{fig:async}
\end{figure}

\noindent\textbf{Asynchronous Prediction and Execution.}
Despite the efficiency gains from KV cache and partial denoising, autoregressive prediction still incurs non-negligible latency that can violate real-time control requirements.
To address this, we introduce an asynchronous inference strategy that pipelines action prediction with execution, effectively hiding prediction latency.
We illustrate the difference between synchronous and asynchronous inference in \cref{fig:async}.

The key insight is to overlap computation with execution (\cref{fig:async}B): While the robot executes the current action chunk $a_t$, the model simultaneously predicts the subsequent action chunk $a_{t+1}$ conditioned on the most recent real observation $z_{t-1}$ (received after the execution of $a_{t-1}$).
For simplicity, we use $z_t$ to denote latent observations (ignoring the video VAE compression) instead of $o_t$ in this section.
We discard all history data before timestamp $t-1$ and use the hat notation $\hat{~}$ to mark predicted visual content. Consequently, the model's active context is limited to the executed action chunk $a_{t-1}$, the recent ground-truth observation $z_{t-1}$, the currently executing action $a_t$, and its corresponding visual forecast $\hat{z}_t$. A naive auto-regressive implementation (\cref{fig:async}B-1) is to store these tokens into the KV cache and predict $\hat{z}_{t+1}$. However, we observed that such a design frequently leads to open-loop degradation and trajectory drift. Because the video generative model inherently favors temporal smoothness, it tends to "continue" the hallucinated video $\hat{z}_t$ while ignoring the critical physical feedback provided by the real observation $z_{t-1}$, eventually causing the model to lose its capacity to react to the environment.

To mitigate this, we introduce a Forward Dynamics Model (FDM) grounded step into our inference pipeline (\cref{fig:async}B-2). Instead of relying on stale forecasts, we replace it by executing a forward dynamics pass: the model uses the recent feedback $z_{t-1}$ and "imagines" the resulting visual state $z_{t}$ after applying action $a_t$. By caching this feedback-grounded prediction instead of a stale forecast, we force the model to re-align with environmental feedback before predicting $z_{t+1}$. This design enhances our asynchronous algorithm into a robust closed-loop system, enabling the robot to effectively perceive and react to real-world changes. 

Algorithm~\ref{alg:async_inference} formalizes this asynchronous pipeline. During post training, we additionally incorporate a forward dynamics prediction loss:
\begin{equation}
\mathcal{L}_{\mathrm{fdm}} = \mathbb{E}_{t, s, \hat{z}_{t+1}, \epsilon} \left[ \| v_\psi(\tilde{z}_{t+1}, s , z_t, a_t, \tilde{z}_{<t}, \hat{a}_{<t} | c) - \dot{z}_{t+1}^{(s)} \|^2 \right],
\end{equation}

\begin{algorithm}[t]
\caption{Asynchronous Inference and Execution}
\label{alg:async_inference}
\begin{algorithmic}[1] 
\Require Initial observation $o_0$, chunk size $K$, KV cache $\mathcal{C}$

\State $z_0 \gets E(o_0)$; $\mathcal{C} \gets \{z_0\}$
\State $\tilde{z}_{1:K}, a_{0:K-1} \gets \Func{Predict}(\mathcal{C})$ \Comment{Cold Start}
\State $\Var{ObsQueue} \gets \emptyset$ \Comment{Thread-safe queue for incoming real observations}
\State $t \gets 0$

\State \textbf{loop}

    \State \hspace{\algorithmicindent} \textbf{parallel}:
    
    \Statex \hspace{\algorithmicindent}\IndBranch \textbf{Branch A: Robot Execution}
    
    \State \hspace{\algorithmicindent}\IndCode \tikzmark{markA_start}\textbf{async} $\Func{Executor}(a_{t:t+K-1}, \Var{ObsQueue})$ \Comment{Execute pre-computed actions}\tikzmark{markA_end}
    
    \Statex \hspace{\algorithmicindent}\IndBranch \textbf{Branch B: Inference with FDM Grounding}
    
    \State \hspace{\algorithmicindent}\IndCode \tikzmark{markB_start}\textbf{if} $t > 0$ \textbf{then}
    
        \Statex \hspace{\algorithmicindent}\IndInner $o_{t-K+1:t} \gets \Var{ObsQueue.dequeue}()$ \Comment{Get real observation}
        \Statex \hspace{\algorithmicindent}\IndInner $z_{t-K+1:t} \gets E(o_{t-K+1:t})$
        \Statex \hspace{\algorithmicindent}\IndInner $\mathcal{C} \gets \mathcal{C} \cup \{z_{t-K+1,t}, a_{t-K:t-1}\}$ \Comment{Cache feedback}
        
    \Statex \hspace{\algorithmicindent}\IndCode \textbf{end if}
    
    \Statex \hspace{\algorithmicindent}\IndCode $\mathcal{C}_{\text{tmp}} \gets \mathcal{C} \cup \{a_{t:t+K-1}\}$ \Comment{Cache action being executed}
    \Statex \hspace{\algorithmicindent}\IndCode $z_{t+1:t+K} \gets \Func{FDM}(\mathcal{C}_{\text{tmp}})$ \Comment{Imagine visual outcome}
    \Statex \hspace{\algorithmicindent}\IndCode $\mathcal{C}_{\text{tmp}} \gets \mathcal{C}_{\text{tmp}} \cup \{z_{t+1:t+K}\}$ \Comment{Update cache}
    \Statex \hspace{\algorithmicindent}\IndCode $\tilde{z}_{t+K+1:t+2K}, a_{t+K:t+2K-1} \gets \Func{Predict}(\mathcal{C}_{\text{tmp}})$
    
    \Statex \hspace{\algorithmicindent}\IndCode $t \gets t + K$\tikzmark{markB_end}

\State \textbf{end loop}

\end{algorithmic}

\DrawScopeLine{markA_start}{markA_end}{0.8em}{2.2ex}{0.8ex}
\DrawScopeLine{markB_start}{markB_end}{0.8em}{2.2ex}{0.8ex}

\end{algorithm}

\section{Experiments}\label{sec:exp}

\subsection{Dataset Curation and Preprocessing}

We curate a large-scale training corpus by aggregating existing public robot manipulation datasets.
All datasets undergo preprocessing to ensure consistency in data format and annotation quality, and are split into 90\% training and 10\% validation per dataset to monitor training dynamics.

\paragraph{Unified Action Representation.}
To achieve cross-embodiment generalization, we define a universal action interface to adapt to different datasets.
We use a dual-arm representation where each robotic arm is characterized by both end-effector pose (EEF) and joint angles.
The end-effector pose consists of XYZ coordinates and a rotation quaternion (7 dimensions).
For joint angles, we support a maximum of 7 degrees of freedom for single-arm embodiments; if a robot has fewer than 7 joint dimensions, we pad the missing dimensions with zeros to maintain a unified 7-dimensional representation.
Each arm also has one gripper action dimension.
Therefore, the total action dimensionality for dual-arm systems is: $7_{\text{EEF}} + 7_{\text{joints}} + 1_{\text{gripper}}$ per arm, resulting in $(7 + 7 + 1) \times 2 = 30$ dimensions.

\paragraph{Training Data Composition.}
We aggregate data from six sources spanning diverse embodiments, environments, and task categories:
\begin{itemize}[leftmargin=*, itemsep=2pt, topsep=2pt]
    \item \textbf{Agibot}~\cite{agibot}: Large-scale dataset with diverse manipulation tasks from mobile manipulators.%
    \item \textbf{RoboMind}~\cite{robomind}: Multi-embodiment manipulation demonstrations.%
    \item \textbf{InternData-A1}~\cite{interndata}: Large-scale simulation dataset for sim-to-real transfer.%
    \item \textbf{OXE}~\cite{oxe}: Multi-embodiment dataset; we use the OpenVLA subset.%
    \item \textbf{UMI Data}~\cite{umi,mvumi,vitamin,fastumi,datascalinglaws,maniwav}: Human demonstration dataset collected via universal manipulation interface\footnote{\url{https://umi-data.github.io/}}, excluding DexUMI. %
    \item \textbf{RoboCOIN}~\cite{robocoin}: Cross-embodiment bimanual robotics data.%
\end{itemize}
In total, our training corpus comprises approximately \textbf{16K} hours of robot manipulation data across diverse tasks and environments, including internally collected demonstrations.

\subsection{Implementation \& Training Details}

\paragraph{Implementation Details.}
We use Wan2.2-5B as the backbone for the video stream, with hidden dimension $d_v = 3072$ and 30 transformer layers.
The action stream shares the same depth but uses a reduced hidden dimension $d_a = 768$ ($4\times$ smaller), resulting in approximately 350M additional parameters and a total model size of 5.3B parameters.
Both streams employ RoPE positional encoding and are connected via the MoT architecture described in \S\ref{sec:architecture}.
We adopt the Wan2.2 causal VAE for tokenization with a $4\times 16\times 16$ (temporal $\times$ height $\times$ width) compression ratio, combined with a patchify operation that further reduces spatial dimensions by 2. The encoded views are concatenated along the width dimension, resulting in a total of $N = 192$ spatial tokens per frame.
The action encoder $\phi$ and decoder are implemented as single-layer MLPs with hidden dimension 256.
We normalize actions using per-dimension quantile normalization statistics computed from the training set.
Task instructions are encoded using a frozen T5 text encoder~\cite{raffel2020exploring} and injected via cross-attention.
During training, chunk size $K$ is randomly sampled from $[1, 4]$.

For inference, we use Euler solver with 3 steps for video tokens (integrating to $s = 0.6$) and 10 steps for action tokens (integrating to $s = 1.0$).
Video CFG scale is set to 5.0, while action CFG scale is set to 1.0.
During training, noise augmentation is applied with probability $p = 0.5$ and $s_{\text{aug}} \sim \text{Uniform}[0.5, 1.0]$.
Following LLM practices, we pack multiple episodes into long sequences (up to 10K tokens) with attention masks.

\paragraph{Pre-Training Details.}
We pretrain \method on the curated dataset for 1.4T tokens.
We use the AdamW optimizer with peak learning rate $1 \times 10^{-4}$, weight decay 0.01, and cosine annealing schedule with linear warmup.
Training is conducted in bfloat16 mixed precision with gradient clipping at 2.0.
We apply classifier-free guidance with text dropout rate 0.1.
The loss weight $\lambda$ for inverse dynamics is set to 1.
The dataset is sampled uniformly across all sources to ensure balanced learning.
We monitor convergence using flow matching loss on the validation set.
We use uniform SNR sampler for video model.
For both video and action model, we use a uniform SNR sampler.

\paragraph{Post-Training Details.}
While the pretrained model exhibits zero-shot generalization to seen embodiments, adapting to novel robot platforms requires a small amount of task-specific data.
We find that post-training with as few as 50 demonstrations is sufficient for effective deployment.
We use a reduced learning rate of $1 \times 10^{-5}$ and train for 3K steps, which yields robust performance.
Alternatively, a higher learning rate of $1 \times 10^{-4}$ with 1K steps also produces reasonable results, though slightly inferior, offering a faster adaptation option when computational resources are limited.

\begin{figure}[t]
    \centering
    \includegraphics[width=0.85\textwidth]{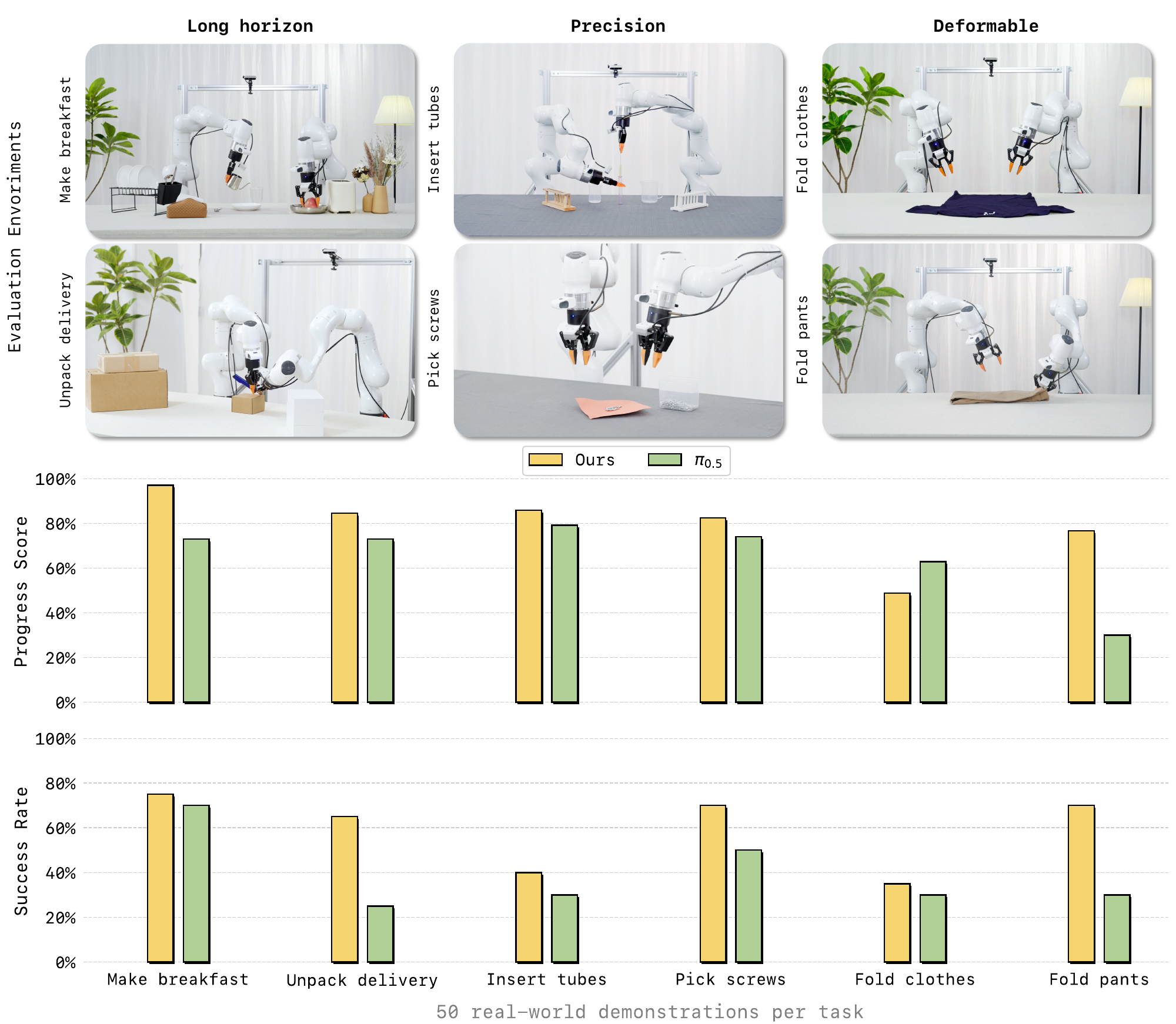}
    \caption{\textbf{Real-world deployment results.} 
 We evaluate \method on six manipulation tasks across three categories: long-horizon tasks (Make Breakfast, Pick Screws), precision tasks (Insert Tubes, Unpack Delivery), and deformable \& articulated object manipulation (Fold Clothes, Fold Pants). Our method achieves state-of-the-art performance on both metrics.}
    \label{fig:realworld}
\end{figure}

\begin{figure}
    \centering
    \includegraphics[width=1\linewidth]{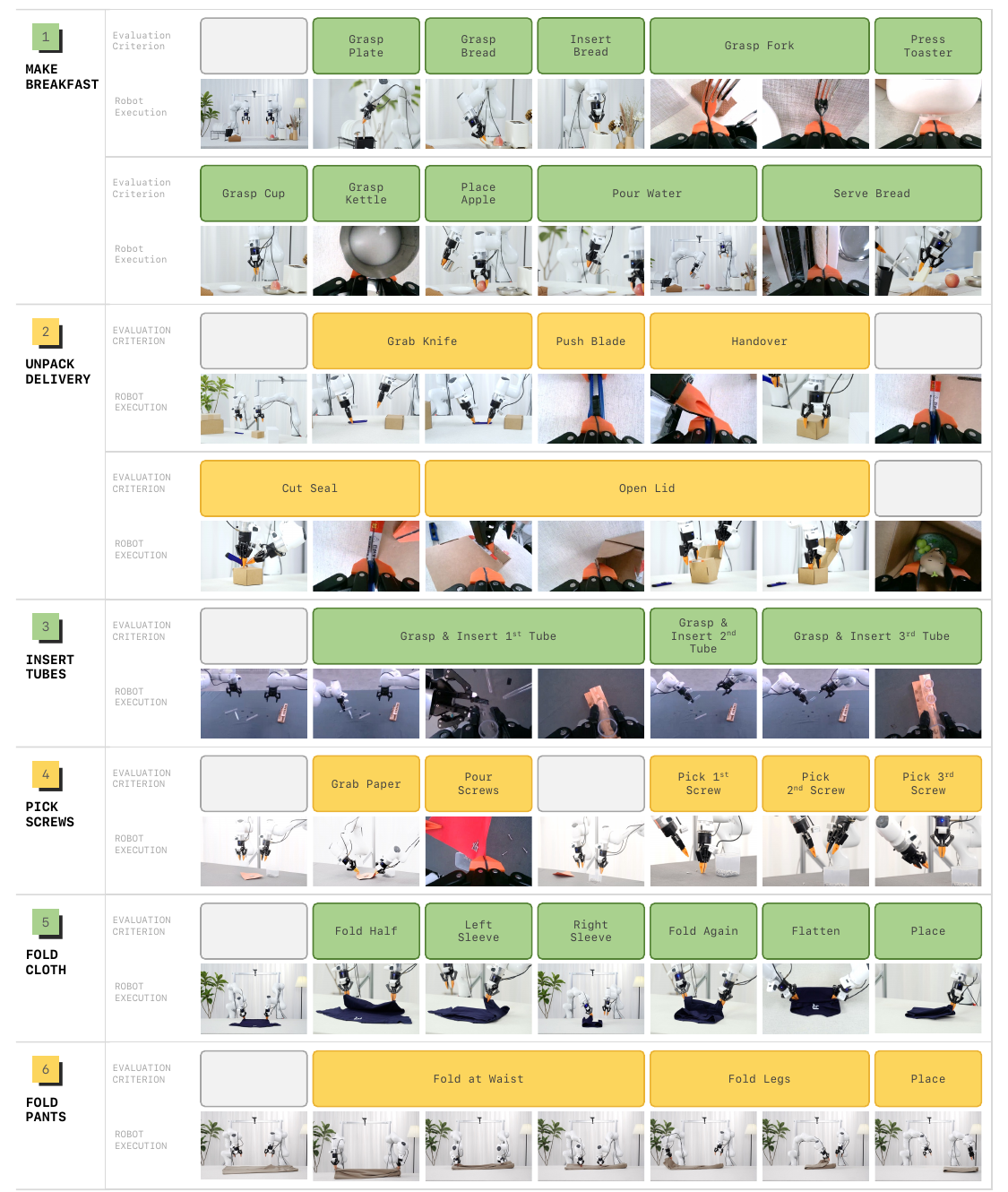}
    \caption{\textbf{Detailed task progressions and key execution steps of the six real-world tasks.} Each task involves a sequence of manipulation primitives, with scoring criteria detailed in Tables \ref{tab:realworld_detailed_make_breakfast} through \ref{tab:realworld_detailed_fold_clothes}.}
    \label{fig:demo_fig}
\end{figure}

\subsection{Main Results}

\subsubsection{Real-world Deployment}
\paragraph{Experimental Setup.} To validate the real-world effectiveness of \method, we deploy our model on a physical robot platform and evaluate across six diverse manipulation tasks spanning three challenging categories.
(1) \textbf{\textit{Long-horizon Tasks:}} We evaluate on \textit{Make Breakfast} and \textit{Unpack Delivery}, which require sequential multi-step reasoning and sustained task execution over extended time horizons.
(2) \textbf{\textit{Precision Tasks:}} We test on \textit{Insert Tubes} and \textit{Pick Screws}, demanding accurate positioning and fine-grained motor control for successful completion.
(3) \textbf{\textit{Deformable Objects:}} We include \textit{Fold Clothes} and \textit{Fold Pants}, which involve manipulating non-rigid materials that present unique control challenges.
The detailed task procedures are summarized in \cref{fig:demo_fig}. 
These tasks are only collected with \textbf{50 real-world demos} for model training.
 We finetune the model for 500 steps with a learning rate of $1 \times 10^{-4}$ and a sequence length of 150,000.

\paragraph{Results.}  As shown in \cref{fig:realworld}, \method consistently achieves state-of-the-art performance across all six tasks and both evaluation metrics (success rate and progress score), substantially outperforming strong baseline $\pi_{0.5}$.

We highlight several key observations that validate our design choices:
(1) The superior performance on \textit{long-horizon tasks} demonstrates that our video-action world model possesses strong temporal memory capabilities. By jointly modeling video and action sequences, the model effectively maintains task context over extended horizons, enabling coherent multi-step reasoning without losing track of intermediate goals.
(2) The strong results on \textit{precision tasks} validate the effectiveness of our unified latent space design. By aligning video and action representations within a shared embedding space, our model achieves tighter coupling between visual perception and motor control, resulting in more accurate and fine-grained action predictions.
(3) The robust performance on \textit{deformable objects} highlights the value of video generation as implicit guidance. The generated video futures provide rich predictive signals about object dynamics and state transitions, which inform the action model to produce more physically plausible manipulation trajectories for challenging non-rigid materials.

These results collectively demonstrate that our video-action world model effectively transfers to real-world deployment, exhibiting robust performance across diverse manipulation scenarios.

\begin{table}[b]
\centering
\caption{\raggedright \textbf{Evaluation on RoboTwin 2.0 Simulation (Easy vs Hard, 50 tasks)}. RoboTwin 2.0 is a challenging bimanual manipulation benchmark requiring coordinated dual-arm control. Easy uses fixed initial configurations while Hard involves randomized object poses and scene layouts. $^{*}$ Results for X-VLA are adopted from Motus~\citep{motus}. Improvements in parentheses indicate gains over the second-best method (underlined).}
\label{tab:robotwin}
\vspace{-5pt}
\small
\begin{tabular}{l *{5}{cc}}
\toprule
& \multicolumn{2}{c}{X-VLA$^{*}$~\citep{xvla}}
& \multicolumn{2}{c}{$\pi_0$~\citep{pi0}}
& \multicolumn{2}{c}{$\pi_{0.5}$~\citep{pi2025pi05}}
& \multicolumn{2}{c}{Motus~\citep{motus}}
& \multicolumn{2}{c}{\method (Ours)} \\
\cmidrule(lr){2-3}\cmidrule(lr){4-5}\cmidrule(lr){6-7}\cmidrule(lr){8-9}\cmidrule(lr){10-11}
\textbf{Metric} 
& Easy & Hard
& Easy & Hard
& Easy & Hard
& Easy & Hard
& Easy & Hard \\
\midrule
$\text{Average}_{\text{ Horizon = 1}}$ & 81.6 & 82.5 & 66.5 & 61.6 & 85.1 & 80.2 & \underline{91.0} & \underline{90.6} & \textbf{94.18} {\scriptsize (+3.2)} & \textbf{93.56} {\scriptsize (+3.0)} \\
$\text{Average}_{\text{ Horizon = 2}}$ & 59.3 & 55.9 & 66.1 & 54.7 & 79.3 & 73.0 & \underline{85.2} & \underline{80.9} & \textbf{90.35} {\scriptsize (+5.2)} & \textbf{86.95} {\scriptsize (+6.1)} \\
$\text{Average}_{\text{ Horizon = 3}}$ & 61.2 & 66.0 & 61.6 & 50.2 & 78.6 & 67.4 & \underline{85.0} & \underline{84.2} & \textbf{93.22} {\scriptsize (+8.2)} & \textbf{93.28} {\scriptsize (+9.1)} \\
\midrule
$\text{Average}_{\text{ 50 Tasks}}$   & 72.9 & 72.8 & 65.9 & 58.4 & 82.7 & 76.8 & \underline{88.7} & \underline{87.0} & \textbf{92.93} {\scriptsize (+4.2)} & \textbf{91.55} {\scriptsize (+4.6)} \\
\bottomrule
\end{tabular}
\label{tab:avg-clean-aug}
\end{table}

\subsubsection{Simulation Evaluation}
\paragraph{Experimental Setup.} We evaluate \method on two widely-used simulation benchmarks: RoboTwin 2.0~\cite{robotwin} and LIBERO~\cite{libero}, covering diverse manipulation tasks across different robot embodiments.

(1) In RoboTwin 2.0, we adopt a multi-task training setup~\cite{motus} where all models are trained on 2,500 demonstrations collected in clean scenes (50 per task) plus 25,000 demonstrations from heavily randomized scenes (500 per task). We downsample the original 50 Hz video to 12.5 Hz while maintaining the action frequency at 50Hz. The model is trained for 50K steps with a learning rate of $1\times10^{-5}$.
To facilitate a clearer comparison of performance, we categorize the 50 RoboTwin tasks according to their horizons (e.g., \textit{Place Dual Shoes} has two steps, and \textit{Stack Blocks Three} has three steps). The detailed horizons are listed in~\cref{tab:robotwin_full}.

(2) In LIBERO, we train our model on four LIBERO suites: LIBERO-Spatial, LIBERO-Object, LIBERO-Goal, and LIBERO-Long. Each suite contains 10 tasks with 50 demonstrations per task (500 total). Following OpenVLA~\cite{kim2024}, we filter unsuccessful demonstrations before training. The model is finetuned for 4K steps with a learning rate of $1 \times 10^{-5}$ and a sequence length of $1 \times 10^{5}$.  Specifically, we report the average success rate over three random seeds, with each seed comprising 500 evaluation trials (totally $3 \times 500 = 1500$) for every task suite.

\paragraph{Results}
RoboTwin 2.0 is a challenging bimanual manipulation benchmark featuring over 50 tasks that require coordinated dual-arm control.
Unlike single-arm benchmarks, RoboTwin tasks demand precise synchronization between two manipulators, making it significantly more difficult for policy learning.
We evaluate under both \textit{Easy} (fixed initial configurations) and \textit{Hard} (varied object poses and scene layouts) settings.
As shown in \cref{tab:avg-clean-aug}, \method achieves an average success rate of 92.9\% (Easy) and 91.6\% (Hard), substantially outperforming prior methods including $\pi_0$, $\pi_{0.5}$, X-VLA, and Motus.
Notably, the improvement becomes more pronounced for longer-horizon tasks: at Horizon = 3, our method achieves gains of +8.2\% (Easy) and +9.1\% (Hard) over the second-best approach.
This suggests that our autoregressive mechanism effectively maintains long-range temporal memory, enabling more robust performance as task complexity increases.

We further evaluate on LIBERO benchmark (\cref{tab:main_results}).
On LIBERO, we obtain an average success rate of 98.5\%, with particularly strong performance on LIBERO-Long (98.5\%).
These results establish new state-of-the-art performance in average success rates among foundational VLAs, demonstrating the effectiveness of our video-action world model for generalist robot control.

\begin{table}[t]
\centering
\caption{\raggedright \textbf{Evaluation on LIBERO benchmarks}. LIBERO tests manipulation across four task suites: Spatial, Object, Goal, and Long-horizon. Our method achieves new state-of-the-art on LIBERO-Object (99.6\%), LIBERO-Long (98.5\%), LIBERO-Spatial (98.5\%), and overall average (98.5\%). Baseline results are adopted from \cite{xvla}.
}
\label{tab:main_results}
\small
\begin{tabular}{l|ccccc}
\toprule
\multirow{2}{*}{\textbf{Methods}} & \multicolumn{5}{c}{\textbf{LIBERO}} \\
& Spatial & Object & Goal & Long & Avg \\
\midrule
Octo \citep{octo2024} & 78.9 & 85.7 & 84.6 & 51.1 & 75.1 \\
Seer \citep{tian2025} & - & - & - & 87.7 & - \\
MoDE \citep{reuss2024} & - & - & - & 94.0 & - \\
SuSIE \citep{black2024b} & - & - & - & 76.3 & - \\
SpatialVLA \citep{qu2025} & 88.2 & 89.9 & 78.6 & 55.5 & 78.1 \\
TraceVLA \citep{zheng2024b} & 84.6 & 85.2 & 75.1 & 54.1 & 74.8 \\
CoT-VLA~\citep{zhao2025cot} & 87.5 & 91.6 & 87.6 & 69.0 & 81.1 \\
ThinkAct \citep{huang2025} & 88.3 & 91.4 & 87.1 & 70.9 & 84.4 \\
SmolVLA \citep{shukor2025} & 93.0 & 94.0 & 91.0 & 77.0 & 88.8 \\
CronusVLA \citep{li2025cronusvla} & 97.3 & 99.6 & 96.9 & 94.0 & 97.0 \\
FLOWER \citep{reuss2025} & 97.1 & 96.7 & 95.6 & 93.5 & 95.7 \\
GR00T-N1 \citep{bjorck2025} & 94.4 & 97.6 & 93.0 & 90.6 & 93.9 \\
$\pi_0$ \citep{pi0} & 96.8 & 98.8 & 95.8 & 85.2 & 94.1 \\
$\pi_0$+FAST \citep{pertsch2025} & 96.4 & 96.8 & 88.6 & 60.2 & 85.5 \\
OpenVLA \citep{kim2024} & 84.7 & 88.4 & 79.2 & 53.7 & 76.5 \\
OpenVLA-OFT \citep{kim2025} & 97.6 & 98.4 & \textbf{97.9} & 94.5 & 97.1 \\
DD-VLA \citep{liang2025} & 97.2 & 98.6 & 97.4 & 92.0 & 96.3 \\
UniVLA \citep{wang2025a} & 95.4 & 98.8 & 93.6 & 94.0 & 95.4 \\
X-VLA~\citep{xvla} & 98.2 & 98.6 & 97.8 & 97.6 & 98.1 \\
\midrule
 \method (Ours) & {$\mathbf{98.5\pm0.3}$} & {$\mathbf{99.6\pm 0.3 }$} & $97.2\pm 0.2 $& {$\mathbf{98.5\pm 0.5}$} & {$\mathbf{98.5}$} \\
\bottomrule
\end{tabular}
\end{table}

\subsection{Ablation}

\noindent\textbf{Asynchronous v.s. synchronous}.
We compare our asynchronous video-action generation with a synchronous baseline on RoboTwin tasks.
As shown in \cref{tab:ablation}, both approaches achieve comparable success rates, but our asynchronous method completes tasks \textbf{2$\times$ faster} by predicting future video and action sequences while executing current actions.
This validates that asynchronous generation maintains task performance while significantly improving inference efficiency.

\noindent\textbf{Pretrained} \method \textbf{v.s. WAN}.
To validate the design choices in our video-action architecture, we conduct a controlled ablation study comparing our pretrained \method model with WAN (Wan2.2-5B) as the initialization baseline for fine-tuning on RoboTwin tasks.
Both models are fine-tuned on the same RoboTwin dataset using identical post-training procedures (50 task-specific demonstrations, learning rate $1 \times 10^{-5}$, 3K steps).

As shown in \cref{tab:ablation}, our pretrained \method model substantially outperforms WAN fine-tuning across both Easy and Hard settings.
Specifically, \method achieves an average success rate of 92.10\% (Easy) and 91.12\% (Hard), while WAN fine-tuning yields significantly lower performance.
This performance gap highlights the effectiveness of our joint video-action pretraining strategy, which endows the model with rich visual-motor priors that facilitate fast adaptation to complex bimanual manipulation tasks.

\noindent\textbf{Action Network Initialization}.
Proper initialization of the action stream is critical for training stability and convergence.
We compare our curated initialization strategy (Section~\ref{sec:architecture}) with naive random initialization.

\begin{table}[t]
\centering
\caption{\textbf{Ablation studies on RoboTwin 2.0 (Easy).} We ablate three design choices: world modeling (AR vs. bidirectional), deployment mode (async vs. sync), and pretraining (Ours vs. WAN).}
\label{tab:ablation}
\small
\begin{tabular}{l l c c c c }
\toprule
\textbf{Ablation} & \textbf{Setting} & \textbf{Easy}$_{\text{all}}$  & \textbf{Easy}$_{\text{ Horizon = 1}}$  & \textbf{Easy}$_{\text{ Horizon = 2}}$  & \textbf{Easy}$_{\text{ Horizon = 3}}$ \\
\midrule
\multirow{1}{*}{Baseline} & \method (Ours) & 92.9 & 94.2 & 90.4 & 93.2  \\
\midrule
\multirow{2}{*}{Deployment} & FDM-grounded Async  & 90.4 & 92.5 & 87.7 & 85.6\\
 & Naive Async & 74.3 & 83.3 & 70.3 & 32.9 \\
\midrule
\multirow{1}{*}{Pretrain}  & WAN & 80.6  & 84.9& 76.3 & 67.6\\
\bottomrule
\end{tabular}
\end{table}

\begin{figure}[t]
    \centering
    \includegraphics[width=0.9\textwidth]{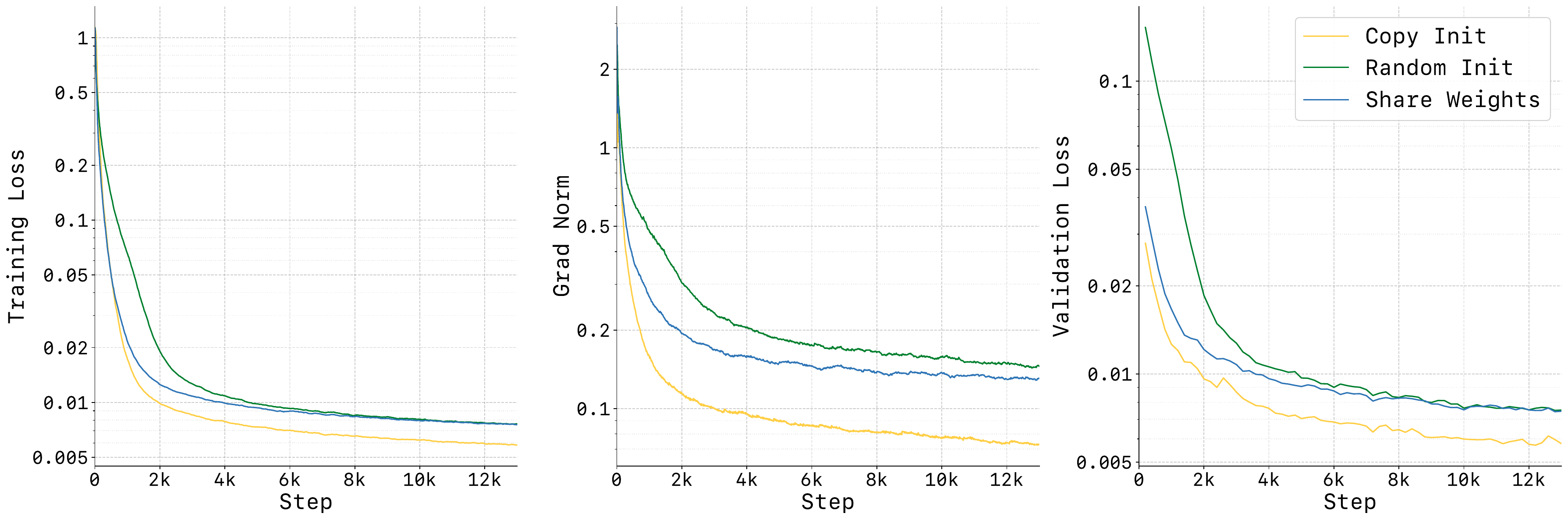}
    \vspace{-5pt}
    \caption{\textbf{Training dynamic comparison between different action network initialization streategy}: Random initialization leads to unstable optimization (high gradient norms) and slow convergence. Although re-using video network weights stabilizes training, the resulting performance is not optimal. Our approach, which initializes by copying pretrained video weights with proper scaling, proves to be the most effective, ensuring smooth training dynamics and faster convergence.
    }
    \label{fig:loss_comparison}
    \vspace{-10pt}
\end{figure}

As shown in \cref{fig:loss_comparison}, random initialization from scratch exhibits volatile training dynamics with significantly slower convergence.
This instability arises because action tokens' output distribution initially diverges dramatically from the video distribution, disrupting the joint attention mechanism in our unified architecture.
In contrast, our curated initialization strategy---where action network weights are initialized by interpolating pretrained video weights with a scaling factor $\alpha = \sqrt{d_v / d_a}$---produces smooth convergence and substantially lower loss.

\subsection{Analysis}

\subsubsection{Sample Efficiency}
\begin{figure}[t]
    \centering
    \includegraphics[width=\textwidth]{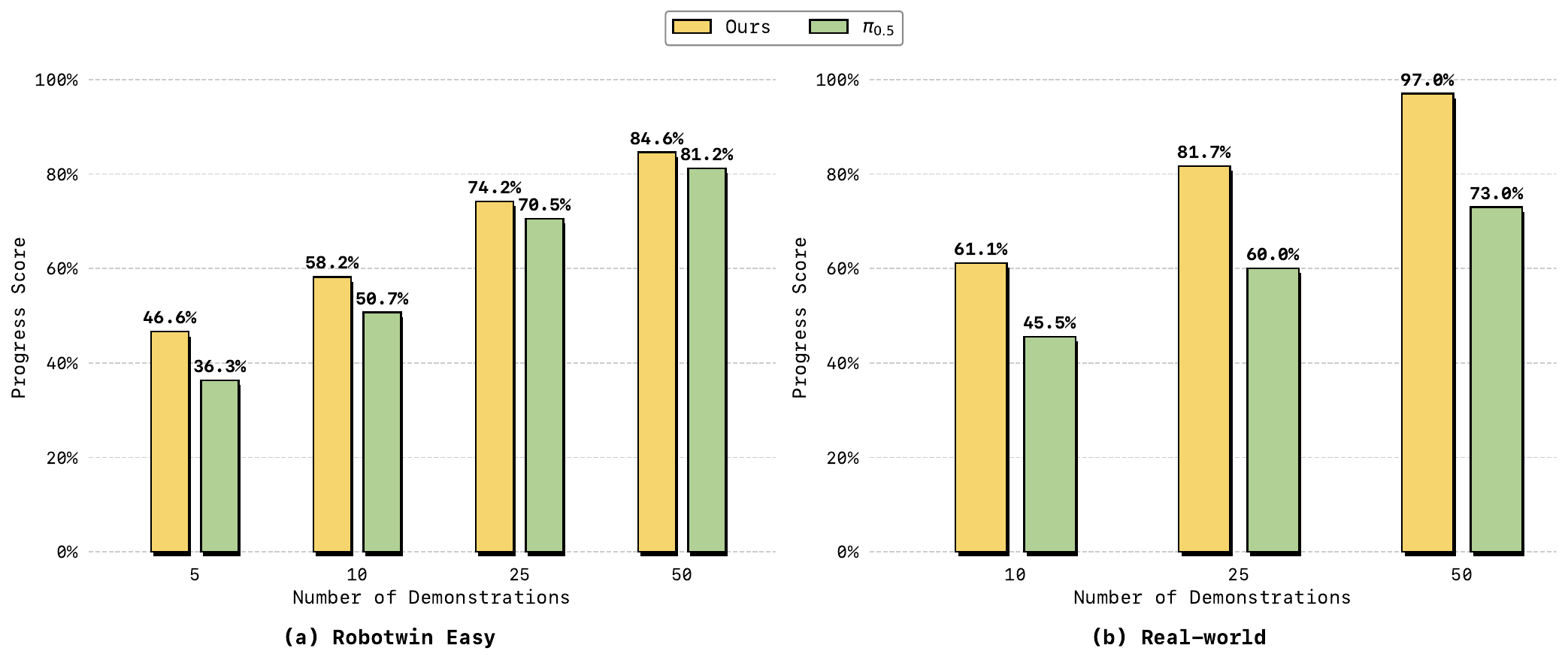}
    \caption{\textbf{Sample efficiency comparison}. \method consistently outperforms $\pi_{0.5}$ across various data regimes on the ``Make Breakfast'' task, demonstrating superior data efficiency in the post-training stage.}
    \label{fig:few_shot}
\end{figure}

We investigate the data efficiency by exploring how \method performs with limited post-training data compared to $\pi_{0.5}$.
We conduct this evaluation on both real-world and simulation settings: the ``Make Breakfast'' long-horizon task and RoboTwin 2.0 Easy benchmarks, allowing us to assess data efficiency across diverse manipulation scenarios.

As shown in \cref{fig:few_shot}, our method consistently outperforms $\pi_{0.5}$ across all data regimes on both real-world and simulation tasks.
In the low-data regime (10 demonstrations), \method achieves 15.6\% higher progress score than $\pi_{0.5}$ on the ``Make Breakfast'' task and 10.3\% higher on RoboTwin 2.0 Easy, demonstrating superior sample efficiency.
These results demonstrate that our method learns more effectively from limited data across diverse manipulation scenarios.

We attribute this superior data efficiency to our video-action world model design.
The jointly pretrained video generation backbone provides rich visual priors about physical dynamics and object interactions, which serve as implicit regularization during post-training.
This allows the action model to leverage the world knowledge encoded in the video stream, effectively reducing the sample complexity required for adapting to new tasks.
In contrast, VLA models like $\pi_{0.5}$ lack explicit modeling of visual dynamics and thus have no structured dynamics priors to guide learning, requiring more demonstrations to learn task-specific behaviors from scratch.

\subsubsection{Temporal Memory}
\begin{figure}[t]
    \centering
    \includegraphics[width=\textwidth]{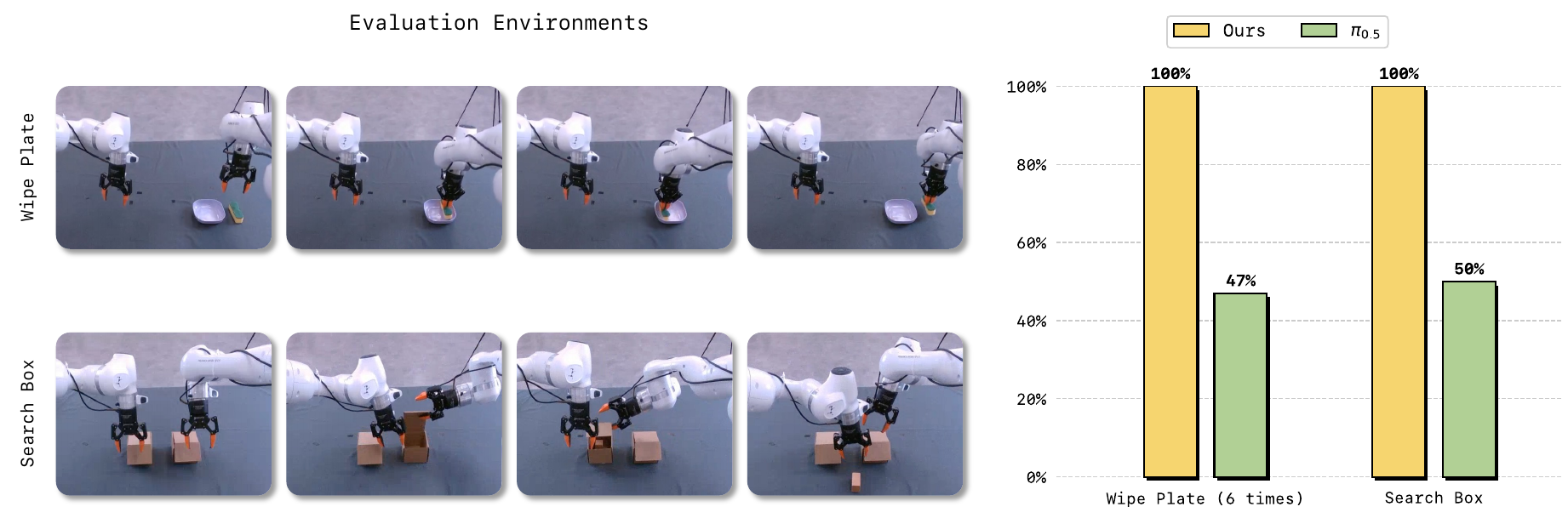}
    \vspace{-5pt}
    \caption{\textbf{Temporal memory evaluation.} Left: Success rates on two memory tasks (Wipe Plate and Search Box). \method significantly outperforms $\pi_{0.5}$ on both tasks, demonstrating superior temporal state tracking ability. Right: Visualization of evaluation environments.}
    \label{fig:analysis}
    \vspace{-10pt}
\end{figure}

We design the following tasks that explicitly require maintaining state information across time to evaluate our model's temporal memory capabilities, as shown in Figure \ref{fig:analysis}.

\begin{enumerate}
    \item \textbf{\textit{Wipe Plate}}---the robot must wipe a plate exactly six times, requiring it to count and remember repeated actions.
    \item \textbf{\textit{Search Box}}---Two boxes (left and right) are in the scene, with only one containing a block. The robot opens them sequentially from right to left. In data collection, the block is equally likely to be in either box; at test time, it is always in the left box. Without memory, after finding the right box empty, the model has a 50\% chance of re-opening it. With memory, it proceeds to search the left box.
\end{enumerate}

As shown in \cref{fig:analysis}(a), \method substantially outperforms $\pi_{0.5}$ on both memory tasks.
We attribute this to the autoregressive nature of our world model: during training, teacher forcing conditions predictions on full history; during inference, KV-cache naturally preserves all historical information for persistent memory.

\subsubsection{Generalization}
We evaluate generalization along two axes:

\begin{enumerate}
    \item \textbf{\textit{Novel Object Generalization}}---trained on pick-and-place with a single object, tested on different objects with varying shapes and textures;
    \item \textbf{\textit{Spatial Generalization}}---trained with fixed object positions in a localized region(denoted as in-distribution~(ID)), tested on random placements especially in out-of-distribution~(OOD) regions.
\end{enumerate}

\begin{figure}[t]
    \centering
    \includegraphics[width=\textwidth]{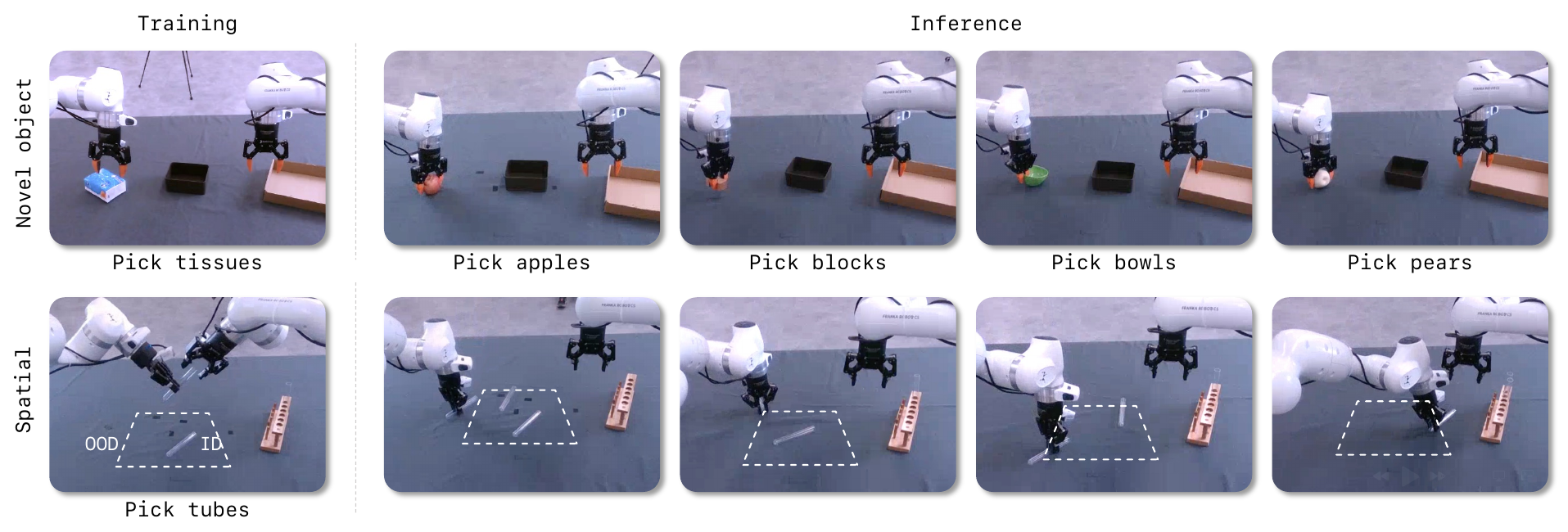}
    \caption{\textbf{Novel object and spatial generalization.} \method successfully generalizes to objects with varying shapes, textures, and positions.}
    \label{fig:generalization}
\end{figure}

As shown in \cref{fig:generalization}, our method demonstrates a stronger generalization in both both novel object and the out-of-distribution position. The world model learns transferable visual representations through video prediction, capturing object-agnostic physical priors that transfer to novel scenarios.

\section{Related Work}
\label{sec:related}

\noindent \textbf{Vision-Language-Action Policies}.
Recent advancements in Embodied AI have witnessed a paradigm shift toward large-scale Vision-Language-Action (VLA) policies. 
By leveraging web-scale knowledge and diverse robot demonstrations, models such as $\pi_{0.5}$~\cite{pi2025pi05}, GR-3~\cite{gr3}, and GR00T-N1~\cite{bjorck2025} achieve remarkable generalizability across various manipulation tasks without relying on hand-crafted rules, modular priors, or restricted action abstractions, enabling a more direct and expressive end-to-end mapping from perception to control.
These policies typically employ pre-trained Vision-Language Models (VLMs) as foundational backbones \cite{pi2025pi05,pi0,gr3,bjorck2025,rt2,kim2024,yang2025_1,xvla}, which provide superior cross-modal understanding and more generalizable action distributions compared to task-specific imitation policies like ACT \cite{act} or Diffusion Policy \cite{diffusionpolicy}.
Efforts have been further devoted to improving the deployability through lightweight backbones \cite{shukor2025,liu2025b,reuss2025}, efficient tokenization \cite{pertsch2025}, real-time inference \cite{rtc,black2025training,tang2025vlash}, or fine-tuning schemes \cite{kim2025,li2025simplevla,jing2025mixture}.
However, despite their prowess in semantic reasoning, a fundamental limitation persists: the pre-training objectives and data distributions of standard VLMs largely overlook the fine-grained system dynamics and low-level trajectories essential for precision manipulation. 
While supervised fine-tuning on expensively collected large-scale robot datasets allows these models to approximate the marginal action distribution \cite{droid,oxe,barreiros2025careful}, they remain deficient in capturing the underlying transition dynamics---specifically, how the physical state of the environment should evolve and will evolve.
Furthermore, most current VLA methods formulate control as a purely reactive mapping from instantaneous observations to actions. This approach inherently fails to account for the historical context necessary to resolve ambiguities in non-Markovian environments. Additionally, the static image-text pre-training inherent in VLMs fails to instill essential temporal priors. Even when augmented with memory modules~\cite{sridhar2025memer, li2025cronusvla,shi2025a}, such models remain unable to reason about the causal and sequential nature of physical interactions.
To address these shortcomings, recent research has pivoted towards generalist robot policies grounded in world models and generative video modeling \cite{onexwm, uva, uwm, shen2025videovla, motus}. 
However, these methods typically generate predictions with bidirectional attention, which violates the causal structure of physical dynamics and lacks persistent long-term memory across the full execution history. Our \method unifies autoregressive video prediction with action decoding under a strict causal temporal structure, where each prediction conditions exclusively on past observations and actions. By maintaining a persistent KV cache over the complete interaction history, \method ensures long-range temporal consistency and allows the policy to synchronize physical execution with the predicted visual evolution of the environment.

\noindent \textbf{World Models for Robotic Control}.
Inspired by human reliance on intuitive physics to anticipate environmental changes, world models aim to facilitate effective planning by predicting future dynamics. Existing approaches are generally categorized into three groups based on their state representations. The first category operates in latent space \cite{watter2015embed, li2019propagation, li2024deformnet, shen2024action}, encoding task-relevant features into compact vectors to predict evolution via probabilistic \cite{li2024deformnet, wu2023daydreamer, lusch2018deep} or deterministic methods \cite{xu2019densephysnet, shen2024action}. The second category utilizes 3D point clouds \cite{shi2024robocraft, wang2023dynamic, sulsky1995application}, leveraging Graph Neural Networks (GNNs) to predict geometric evolution \cite{zhang2024adaptigraph, zhang2025particle}, which is particularly effective for manipulating deformable objects \cite{zhang2025particle, wang2023dynamic}. The third category focuses on 2D pixel space, directly predicting future keyframes or video sequences \cite{zhou2025act2goal, du2023learning, kim2026cosmos, zhou2024robodreamer}. Our work aligns with this third category. Within this domain, approaches range from co-training with video generation for representation learning \cite{uva, uwm, bu2025learning} to serving as simulators for policy learning or evaluation \cite{team2025evaluating}. Our research specifically targets methods that predict future frames during execution to condition action generation. However, prior video-conditioned methods predominantly rely on open-loop generation \cite{du2023learning, zhou2025act2goal}, presenting two significant challenges. First, the misalignment between generated videos and real-world dynamics, coupled with cumulative drift from execution errors, often leads to suboptimal performance. Second, the computational intensity of video generation imposes high latency, severely hindering real-time inference. Our method leverages KV Cache and causal masking to continuously update the model's memory with real-world observations. This effectively transitions the system to a closed-loop control mechanism, mitigating error accumulation in long-horizon tasks. Furthermore, we introduce a partial denoising strategy, enabling action generation from intermediate representations without waiting for fully denoised frames.

\section{Conclusion}\label{sec:conclusion}

We present \method, an autoregressive diffusion framework that unifies video dynamics prediction and action inference for robotic manipulation.
By interleaving video and action tokens within a Mixture-of-Transformers architecture, our model captures the causal structure of physical interactions while enabling closed-loop control through continuous integration of real-world observations.
Extensive evaluation demonstrates strong performance across simulation benchmarks (92.0\% on RoboTwin 2.0, 98.5\% on LIBERO) and real-world deployment, achieving over 20\% improvement on challenging tasks compared to $\pi_{0.5}$ with only 50 demonstrations for adaptation.
These results suggest that autoregressive video-action world modeling provides a principled foundation for learning generalizable manipulation policies, offering a compelling alternative to reactive VLA paradigms.

\noindent\textbf{Future Work.}
Future directions include developing more efficient video compression schemes to reduce computational overhead, and incorporating multi-modal sensory inputs (tactile, force, audio) for more robust manipulation in tasks with complex contact dynamics.

\noindent\textbf{Acknowledgment.}
We thank Kecheng Zheng for insightful discussions and Wei Wu for valuable assistance with dataset preparation.
We also thank Fangyi Xu and Yishu Shen for their help with the post-training data collection.

{
\small
\bibliographystyle{plain}
\bibliography{ref.bib}
}

\appendix
\renewcommand\thesection{\Alph{section}}
\renewcommand\thefigure{S\arabic{figure}}
\renewcommand\thetable{S\arabic{table}}
\renewcommand\theequation{S\arabic{equation}}
\setcounter{figure}{0}
\setcounter{table}{0}
\setcounter{equation}{0}

\section*{Appendix}

\begin{table}[t]
\centering
\caption{\raggedright \textit{Evaluation on RoboTwin 2.0 Simulation (Easy vs Hard, 50 tasks)}. RoboTwin 2.0 is a challenging bimanual manipulation benchmark requiring coordinated dual-arm control. Easy uses fixed initial configurations while Hard involves randomized object poses and scene layouts.}
\label{tab:robotwin_full}
\vspace{-5pt}
\small
\setlength{\tabcolsep}{3pt}
\resizebox{0.8\textwidth}{!}{%
\begin{tabular}{l c cc cc cc cc cc}
\toprule
\multirow{2}{*}{\textbf{Simulation Task}} & \multirow{2}{*}{\textbf{Horizon}} &
\multicolumn{2}{c}{\textbf{Ours}} & \multicolumn{2}{c}{$\pi_0$~\cite{pi0}} & \multicolumn{2}{c}{$\pi_{0.5}$~\cite{pi0}} & \multicolumn{2}{c}{X-VLA~\cite{xvla}} & \multicolumn{2}{c}{Motus~\cite{motus}} \\
\cmidrule(lr){3-4} \cmidrule(lr){5-6} \cmidrule(lr){7-8} \cmidrule(lr){9-10} \cmidrule(lr){11-12}
 &  & Easy & Hard & Easy & Hard & Easy & Hard & Easy & Hard & Easy & Hard \\
\midrule
\textit{Adjust Bottle} & 1 & 90\% & 94\% & 99\% & 95\% & \textbf{100\%} & \textbf{99\%} & \textbf{100\%} & \textbf{99\%} & 89\% & 93\% \\
\textit{Beat Block Hammer} & 1 & \textbf{96\%} & 98\% & 79\% & 84\% & \textbf{96\%} & 93\% & 92\% & 88\% & 95\% & 88\% \\
\textit{Blocks Ranking RGB} & 3 & \textbf{99\%} & 98\% & 80\% & 63\% & 92\% & 85\% & 83\% & 83\% & \textbf{99\%} & 97\% \\
\textit{Blocks Ranking Size} & 3 & \textbf{94\%} & 96\% & 14\% & 5\% & 49\% & 26\% & 67\% & 74\% & 75\% & 63\% \\
\textit{Click Alarmclock} & 1 & 99\% & \textbf{100\%} & 77\% & 68\% & 98\% & 89\% & 99\% & 99\% & \textbf{100\%} & \textbf{100\%} \\
\textit{Click Bell} & 1 & \textbf{100\%} & \textbf{100\%} & 71\% & 48\% & \textbf{99\%} & 66\% & \textbf{100\%} & \textbf{100\%} & \textbf{100\%} & \textbf{100\%} \\
\textit{Dump Bin Bigbin} & 1 & 89\% & 96\% & 88\% & 83\% & 92\% & \textbf{97\%} & 79\% & 77\% & \textbf{95\%} & 91\% \\
\textit{Grab Roller} & 1 & \textbf{100\%} & \textbf{100\%} & 98\% & 94\% & \textbf{100\%} & \textbf{100\%} & \textbf{100\%} & \textbf{100\%} & \textbf{100\%} & \textbf{100\%} \\
\textit{Handover Block} & 2 & 99\% & 78\% & 47\% & 31\% & 66\% & 57\% & 73\% & 37\% & 86\% & 73\% \\
\textit{Handover Mic} & 2 & 94\% & 96\% & \textbf{97\%} & \textbf{97\%} & \textbf{98\%} & \textbf{97\%} & 0\% & 0\% & 78\% & 63\% \\
\textit{Hanging Mug} & 2 & \textbf{40\%} & 28\% & 14\% & 11\% & 18\% & 17\% & 23\% & 27\% & \textbf{38\%} & \textbf{38\%} \\
\textit{Lift Pot} & 1 & \textbf{100\%} & 99\% & 80\% & 72\% & 96\% & 85\% & \textbf{99\%} & \textbf{100\%} & 96\% & \textbf{99\%} \\
\textit{Move Can Pot} & 1 & 94\% & 97\% & \textbf{68\%} & 48\% & 51\% & 55\% & 89\% & 86\% & 34\% & 74\% \\
\textit{Move Pillbottle Pad} & 1 & \textbf{99\%} & 99\% & 67\% & 46\% & 84\% & 61\% & 73\% & 71\% & 93\% & \textbf{96\%} \\
\textit{Move Playingcard Away} & 1 & \textbf{100\%} & 99\% & 74\% & 65\% & 96\% & 84\% & 93\% & \textbf{98\%} & \textbf{100\%} & 96\% \\
\textit{Move Stapler Pad} & 1 & 91\% & 79\% & 41\% & 24\% & 56\% & 42\% & 78\% & 73\% & \textbf{83\%} & \textbf{85\%} \\
\textit{Open Laptop} & 1 & 92\% & 94\% & 71\% & 81\% & 90\% & 96\% & 93\% & \textbf{100\%} & 95\% & 91\% \\
\textit{Open Microwave} & 1 & 82\% & 86\% & 4\% & 32\% & 34\% & \textbf{77\%} & \textbf{79\%} & 71\% & \textbf{95\%} & \textbf{91\%} \\
\textit{Pick Diverse Bottles} & 2 & 89\% & 82\% & 69\% & 31\% & 81\% & 71\% & 58\% & 36\% & 90\% & \textbf{91\%} \\
\textit{Pick Dual Bottles} & 2 & \textbf{100\%} & 99\% & 59\% & 37\% & 93\% & 63\% & 47\% & 36\% & 96\% & 90\% \\
\textit{Place A2B Left} & 1 & \textbf{97\%} & 93\% & 43\% & 47\% & 87\% & 82\% & 48\% & 49\% & 82\% & 79\% \\
\textit{Place A2B Right} & 1 & \textbf{97\%} & 95\% & 39\% & 34\% & 87\% & 84\% & 36\% & 36\% & 90\% & \textbf{87\%} \\
\textit{Place Bread Basket} & 1 & \textbf{97\%} & 95\% & 62\% & 46\% & 77\% & 64\% & 81\% & 71\% & 91\% & \textbf{94\%} \\
\textit{Place Bread Skillet} & 2 & 95\% & 90\% & 66\% & 49\% & \textbf{85\%} & 66\% & 77\% & 67\% & 86\% & \textbf{83\%} \\
\textit{Place Burger Fries} & 2 & 97\% & 95\% & 81\% & 76\% & \textbf{94\%} & 87\% & \textbf{94\%} & \textbf{94\%} & \textbf{98\%} & \textbf{98\%} \\
\textit{Place Can Basket} & 2 & 81\% & 84\% & 55\% & 46\% & \textbf{62\%} & \textbf{62\%} & 49\% & 52\% & \textbf{81\%} & 76\% \\
\textit{Place Cans Plasticbox} & 2 & \textbf{100\%} & 99\% & 63\% & 45\% & 94\% & 84\% & \textbf{97\%} & \textbf{98\%} & \textbf{98\%} & 94\% \\
\textit{Place Container Plate} & 1 & 99\% & 97\% & \textbf{97\%} & 92\% & \textbf{99\%} & 95\% & \textbf{97\%} & 95\% & 98\% & \textbf{99\%} \\
\textit{Place Dual Shoes} & 2 & 94\% & 89\% & 59\% & 51\% & 75\% & 75\% & 79\% & \textbf{88\%} & 93\% & 87\% \\
\textit{Place Empty Cup} & 1 & \textbf{100\%} & \textbf{100\%} & 91\% & 85\% & \textbf{100\%} & 99\% & \textbf{100\%} & 98\% & 99\% & 98\% \\
\textit{Place Fan} & 1 & 99\% & 93\% & 66\% & 71\% & \textbf{87\%} & 85\% & 80\% & 75\% & \textbf{91\%} & 87\% \\
\textit{Place Mouse Pad} & 1 & 93\% & 96\% & 20\% & 20\% & \textbf{60\%} & 39\% & \textbf{70\%} & \textbf{70\%} & 66\% & 68\% \\
\textit{Place Object Basket} & 2 & 91\% & 88\% & 67\% & 70\% & 80\% & 76\% & 44\% & 39\% & 81\% & \textbf{87\%} \\
\textit{Place Object Scale} & 1 & 96\% & 95\% & 57\% & 52\% & \textbf{86\%} & 80\% & 52\% & 74\% & 88\% & 85\% \\
\textit{Place Object Stand} & 1 & \textbf{99\%} & 96\% & 82\% & 68\% & 91\% & 85\% & 86\% & 88\% & \textbf{98\%} & \textbf{97\%} \\
\textit{Place Phone Stand} & 1 & \textbf{97\%} & 97\% & 49\% & 53\% & 81\% & 81\% & \textbf{88\%} & 87\% & 87\% & 86\% \\
\textit{Place Shoe} & 1 & 98\% & 98\% & 76\% & 76\% & 92\% & 93\% & \textbf{96\%} & 95\% & \textbf{99\%} & \textbf{97\%} \\
\textit{Press Stapler} & 1 & 85\% & 82\% & 44\% & 37\% & 87\% & 83\% & \textbf{92\%} & \textbf{98\%} & 93\% & \textbf{98\%} \\
\textit{Put Bottles Dustbin} & 3 & 87\% & 91\% & 65\% & 56\% & \textbf{84\%} & 79\% & 74\% & 77\% & 81\% & 79\% \\
\textit{Put Object Cabinet} & 2 & 85\% & 87\% & 73\% & 60\% & \textbf{80\%} & \textbf{79\%} & 46\% & 48\% & \textbf{88\%} & 71\% \\
\textit{Rotate QRcode} & 1 & \textbf{96\%} & 91\% & 74\% & 70\% & 89\% & 87\% & 34\% & 33\% & \textbf{89\%} & 73\% \\
\textit{Scan Object} & 2 & 96\% & 91\% & 55\% & 42\% & 72\% & 65\% & 14\% & 36\% & 67\% & 66\% \\
\textit{Shake Bottle Horizontally} & 1 & \textbf{100\%} & 99\% & 98\% & 92\% & \textbf{99\%} & \textbf{99\%} & \textbf{100\%} & \textbf{100\%} & \textbf{100\%} & 98\% \\
\textit{Shake Bottle} & 1 & \textbf{100\%} & 97\% & 94\% & 91\% & \textbf{99\%} & 97\% & \textbf{99\%} & \textbf{100\%} & \textbf{100\%} & \textbf{97\%} \\
\textit{Stack Blocks Three} & 3 & 99\% & 98\% & 72\% & 52\% & 91\% & 76\% & 6\% & 10\% & 91\% & \textbf{95\%} \\
\textit{Stack Blocks Two} & 2 & \textbf{100\%} & 98\% & 93\% & 79\% & 97\% & \textbf{100\%} & 92\% & 87\% & \textbf{100\%} & 98\% \\
\textit{Stack Bowls Three} & 3 & 86\% & 83\% & \textbf{77\%} & 75\% & \textbf{77\%} & 71\% & 76\% & \textbf{86\%} & 79\% & \textbf{87\%} \\
\textit{Stack Bowls Two} & 2 & 94\% & 98\% & 94\% & 95\% & 95\% & 96\% & 96\% & 93\% & \textbf{98\%} & \textbf{98\%} \\
\textit{Stamp Seal} & 1 & 96\% & 97\% & 46\% & 33\% & 79\% & 55\% & 76\% & 82\% & 93\% & 92\% \\
\textit{Turn Switch} & 1 & 44\% & 45\% & 41\% & 42\% & \textbf{62\%} & 54\% & 40\% & 61\% & \textbf{84\%} & \textbf{78\%} \\
\midrule
\textbf{Average (\%)} & -- & \textbf{92.93} & \textbf{91.55} & 65.92 & 58.40 & 82.74 & 76.76 & 72.80 & 72.84 & 88.66 & 87.02 \\
\bottomrule
\end{tabular}%
}
\vspace{-5pt}
\end{table}

\section{Real-world Evaluation Details}\label{appendix:real_world_details}

We present detailed evaluation results for all real-world manipulation tasks. Each task is evaluated with 20 trials for both our method and the baseline ($\pi_{0.5}$). To ensure fair comparison, we adopt an alternating evaluation protocol: one trial with $\pi_{0.5}$, followed by one trial with our method, and so on.

For each trial, we record the success status of every intermediate step. If a step requires a retry to succeed, we assign a score of 0.5; if it fails, the score is 0; if it succeeds on the first attempt, the score is 1. A trial is marked as successful only if all steps are completed (i.e., the total score equals the maximum possible score).

We report two metrics:
\begin{itemize}
    \item \textbf{Progress Score (PS)}: The average score across all trials divided by the maximum possible score, expressed as a percentage: $\text{PS} = \frac{\text{Average Progress}}{\text{Max Steps}} \times 100\%$.
    \item \textbf{Success Rate (SR)}: The number of successful trials divided by the total number of trials, expressed as a percentage: $\text{SR} = \frac{\text{\# Successful Trials}}{N} \times 100\%$.
\end{itemize}

We evaluate on six diverse real-world tasks: \textbf{Make Breakfast} (10 steps: preparing a complete breakfast including toasting bread, pouring water, and plating), \textbf{Pick Screws} (5 steps: picking up paper, pouring screws, and inserting three screws), \textbf{Fold Clothes} (6 steps: folding a shirt including sleeves and smoothing), \textbf{Unpack Delivery} (5 steps: opening a package using a utility knife), \textbf{Insert Tubes} (2 categories: grasping and inserting 3 tubes), and \textbf{Fold Pants} (3 steps: folding pants and placing them). These tasks span long-horizon sequential manipulation, precision control, and deformable object handling. The following tables present per-trial results for each task.

\begin{table}[t]
\centering
\caption{Detailed evaluation results for Make Breakfast task (10 steps, max score 10).}
\label{tab:realworld_detailed_make_breakfast}
\scriptsize
\resizebox{\textwidth}{!}{
\begin{tabular}{l|c|cccccccccc|c}
\toprule
\textbf{Trial} & \textbf{Succ.} & \makecell{\textbf{Grasp}\\\textbf{Plate}} & \makecell{\textbf{Grasp}\\\textbf{Bread}} & \makecell{\textbf{Grasp}\\\textbf{Fork}} & \makecell{\textbf{Place}\\\textbf{Bread}} & \makecell{\textbf{Press}\\\textbf{Toaster}} & \makecell{\textbf{Grasp}\\\textbf{Cup}} & \makecell{\textbf{Grasp}\\\textbf{Kettle}} & \textbf{Pour} & \makecell{\textbf{Grasp}\\\textbf{Apple}} & \textbf{Serve} & \textbf{Prog.} \\
\midrule
\multicolumn{13}{c}{\textbf{Ours}} \\
\midrule
1  & 0 & 1 & 1 & 1 & 1 & 1 & 1 & 1 & 0 & 1 & 1 & 9 \\
2  & 1 & 1 & 1 & 1 & 1 & 1 & 1 & 1 & 1 & 1 & 1 & 10 \\
3  & 1 & 1 & 1 & 1 & 1 & 1 & 1 & 1 & 1 & 1 & 1 & 10 \\
4  & 1 & 1 & 1 & 1 & 1 & 1 & 1 & 1 & 1 & 1 & 1 & 10 \\
5  & 1 & 1 & 1 & 1 & 1 & 1 & 1 & 1 & 1 & 1 & 1 & 10 \\
6  & 1 & 1 & 1 & 1 & 1 & 1 & 1 & 1 & 1 & 1 & 1 & 10 \\
7  & 0 & 1 & 1 & 1 & 1 & 1 & 1 & 1 & 1 & 1 & 0 & 9 \\
8  & 1 & 1 & 1 & 1 & 1 & 1 & 1 & 1 & 1 & 0.5 & 1 & 9.5 \\
9  & 0 & 1 & 1 & 1 & 0 & 1 & 1 & 1 & 1 & 1 & 1 & 9 \\
10 & 1 & 1 & 1 & 1 & 1 & 1 & 1 & 1 & 1 & 1 & 1 & 10 \\
11 & 0 & 1 & 1 & 1 & 1 & 1 & 1 & 1 & 0 & 1 & 1 & 9 \\
12 & 1 & 1 & 1 & 1 & 1 & 1 & 1 & 1 & 1 & 1 & 1 & 10 \\
13 & 0 & 1 & 1 & 1 & 0 & 1 & 1 & 1 & 1 & 1 & 1 & 9 \\
14 & 1 & 1 & 1 & 1 & 1 & 1 & 1 & 1 & 1 & 1 & 1 & 10 \\
15 & 1 & 1 & 1 & 1 & 1 & 1 & 1 & 1 & 1 & 1 & 1 & 10 \\
16 & 1 & 1 & 1 & 1 & 1 & 1 & 1 & 1 & 1 & 1 & 1 & 10 \\
17 & 1 & 1 & 1 & 1 & 1 & 1 & 1 & 1 & 1 & 0.5 & 1 & 9.5 \\
18 & 1 & 1 & 1 & 1 & 1 & 1 & 1 & 1 & 1 & 1 & 1 & 10 \\
19 & 1 & 1 & 1 & 1 & 1 & 1 & 1 & 1 & 1 & 1 & 1 & 10 \\
20 & 1 & 1 & 1 & 1 & 1 & 1 & 1 & 1 & 1 & 1 & 1 & 10 \\
\midrule
\textbf{Avg} & \textbf{0.75} & \textbf{1.00} & \textbf{1.00} & \textbf{1.00} & \textbf{0.90} & \textbf{1.00} & \textbf{1.00} & \textbf{1.00} & \textbf{0.90} & \textbf{0.95} & \textbf{0.95} & \textbf{9.70} \\
\textbf{Progress Score} & \multicolumn{12}{c}{\textbf{97.0\%} (= 9.70/10 $\times$ 100\%)} \\
\textbf{Success Rate} & \multicolumn{12}{c}{\textbf{75.0\%} (= 15/20 $\times$ 100\%)} \\
\midrule
\multicolumn{13}{c}{$\boldsymbol{\pi_{0.5}}$} \\
\midrule
1  & 0 & 1 & 1 & 1 & 1 & 0 & 0 & 0 & 0 & 0 & 0 & 4 \\
2  & 1 & 1 & 1 & 1 & 1 & 1 & 1 & 1 & 0 & 1 & 1 & 9 \\
3  & 1 & 1 & 1 & 1 & 1 & 1 & 1 & 1 & 0 & 1 & 1 & 9 \\
4  & 1 & 1 & 1 & 1 & 1 & 0 & 0 & 1 & 1 & 1 & 1 & 8 \\
5  & 0 & 1 & 1 & 1 & 0 & 1 & 0 & 1 & 1 & 0 & 1 & 7 \\
6  & 1 & 1 & 1 & 1 & 1 & 1 & 1 & 1 & 0 & 1 & 1 & 9 \\
7  & 1 & 1 & 1 & 0 & 1 & 1 & 1 & 1 & 0 & 1 & 1 & 8 \\
8  & 0 & 1 & 1 & 1 & 1 & 1 & 1 & 1 & 1 & 1 & 0 & 9 \\
9  & 0 & 1 & 1 & 0 & 1 & 0 & 1 & 1 & 0 & 0 & 0 & 5 \\
10 & 0 & 1 & 1 & 1 & 1 & 0 & 0 & 0 & 0 & 0 & 0 & 4 \\
11 & 1 & 1 & 1 & 1 & 1 & 0 & 0 & 1 & 1 & 1 & 1 & 8 \\
12 & 1 & 1 & 1 & 1 & 0 & 0 & 0 & 1 & 1 & 1 & 1 & 7 \\
13 & 1 & 1 & 1 & 0 & 1 & 0 & 0 & 1 & 1 & 1 & 1 & 7 \\
14 & 1 & 1 & 1 & 1 & 0 & 1 & 0 & 1 & 1 & 1 & 1 & 8 \\
15 & 1 & 1 & 1 & 1 & 1 & 0 & 0 & 1 & 1 & 1 & 1 & 8 \\
16 & 1 & 1 & 1 & 1 & 0 & 1 & 0 & 1 & 1 & 1 & 1 & 8 \\
17 & 1 & 1 & 1 & 1 & 1 & 1 & 0 & 1 & 1 & 1 & 1 & 9 \\
18 & 1 & 1 & 1 & 1 & 1 & 1 & 1 & 0 & 0 & 1 & 1 & 8 \\
19 & 1 & 1 & 1 & 0 & 1 & 0 & 0 & 1 & 0 & 1 & 1 & 6 \\
20 & 0 & 1 & 1 & 1 & 0 & 1 & 0 & 0 & 0 & 1 & 0 & 5 \\
\midrule
\textbf{Avg} & \textbf{0.70} & \textbf{1.00} & \textbf{1.00} & \textbf{0.80} & \textbf{0.75} & \textbf{0.55} & \textbf{0.35} & \textbf{0.80} & \textbf{0.50} & \textbf{0.80} & \textbf{0.75} & \textbf{7.30} \\
\textbf{Progress Score} & \multicolumn{12}{c}{\textbf{73.0\%} (= 7.30/10 $\times$ 100\%)} \\
\textbf{Success Rate} & \multicolumn{12}{c}{\textbf{70.0\%} (= 14/20 $\times$ 100\%)} \\
\bottomrule
\end{tabular}
}
\end{table}

\begin{table}[t]
\centering
\caption{Detailed evaluation results for Pick Screws task (5 steps, max score 5).}
\label{tab:realworld_detailed_pick_screws}
\scriptsize
\begin{tabular}{l|c|ccccc|c}
\toprule
\textbf{Trial} & \textbf{Success} & \textbf{Grab Paper} & \textbf{Pour Screws} & \textbf{Screw 1} & \textbf{Screw 2} & \textbf{Screw 3} & \textbf{Progress} \\
\midrule
\multicolumn{8}{c}{\textbf{Ours}} \\
\midrule
1  & 1 & 1 & 1 & 0.5 & 0.5 & 1 & 4 \\
2  & 0 & 1 & 0 & 1 & 1 & 0.5 & 3.5 \\
3  & 1 & 1 & 1 & 1 & 0.5 & 0.5 & 4 \\
4  & 1 & 1 & 1 & 1 & 1 & 1 & 5 \\
5  & 1 & 1 & 1 & 1 & 1 & 0.5 & 4.5 \\
6  & 1 & 1 & 1 & 0.5 & 1 & 0.5 & 4 \\
7  & 1 & 1 & 1 & 1 & 1 & 1 & 5 \\
8  & 1 & 1 & 1 & 1 & 1 & 1 & 5 \\
9  & 0 & 1 & 0 & 0 & 0 & 0 & 1 \\
10 & 1 & 1 & 1 & 1 & 1 & 1 & 5 \\
11 & 1 & 1 & 1 & 1 & 1 & 1 & 5 \\
12 & 0 & 1 & 0 & 1 & 1 & 1 & 4 \\
13 & 0 & 1 & 0 & 1 & 1 & 1 & 4 \\
14 & 1 & 1 & 1 & 1 & 0.5 & 0.5 & 4 \\
15 & 0 & 1 & 0 & 0 & 0.5 & 1 & 2.5 \\
16 & 1 & 1 & 1 & 1 & 1 & 1 & 5 \\
17 & 1 & 1 & 1 & 1 & 1 & 0.5 & 4.5 \\
18 & 1 & 1 & 1 & 0.5 & 0.5 & 1 & 4 \\
19 & 0 & 1 & 1 & 0 & 1 & 0.5 & 3.5 \\
20 & 1 & 1 & 1 & 1 & 1 & 1 & 5 \\
\midrule
\textbf{Avg} & \textbf{0.70} & \textbf{1.00} & \textbf{0.75} & \textbf{0.78} & \textbf{0.83} & \textbf{0.78} & \textbf{4.13} \\
\textbf{Progress Score} & \multicolumn{7}{c}{\textbf{82.5\%} (= 4.13/5 $\times$ 100\%)} \\
\textbf{Success Rate} & \multicolumn{7}{c}{\textbf{70.0\%} (= 14/20 $\times$ 100\%)} \\
\midrule
\multicolumn{8}{c}{$\boldsymbol{\pi_{0.5}}$} \\
\midrule
1  & 0 & 1 & 0 & 1 & 1 & 1 & 4 \\
2  & 0 & 0.5 & 0 & 1 & 1 & 0.5 & 3 \\
3  & 1 & 1 & 1 & 1 & 1 & 1 & 5 \\
4  & 0 & 1 & 0 & 1 & 0.5 & 0.5 & 3 \\
5  & 1 & 1 & 1 & 0.5 & 1 & 0.5 & 4 \\
6  & 1 & 1 & 1 & 1 & 0.5 & 1 & 4.5 \\
7  & 1 & 1 & 1 & 1 & 1 & 1 & 5 \\
8  & 1 & 1 & 1 & 0.5 & 0.5 & 1 & 4 \\
9  & 0 & 1 & 0 & 0.5 & 0.5 & 0 & 2 \\
10 & 1 & 1 & 1 & 1 & 1 & 1 & 5 \\
11 & 1 & 1 & 1 & 0.5 & 1 & 1 & 4.5 \\
12 & 0 & 1 & 0 & 1 & 1 & 0.5 & 3.5 \\
13 & 0 & 1 & 1 & 1 & 0 & 0.5 & 3.5 \\
14 & 1 & 1 & 1 & 0.5 & 1 & 1 & 4.5 \\
15 & 0 & 1 & 1 & 1 & 1 & 0 & 4 \\
16 & 1 & 1 & 1 & 1 & 1 & 1 & 5 \\
17 & 0 & 0 & 0 & 1 & 0.5 & 0.5 & 2 \\
18 & 0 & 0 & 0 & 0 & 0 & 0 & 0 \\
19 & 0 & 1 & 0 & 1 & 0.5 & 0.5 & 3 \\
20 & 1 & 1 & 1 & 1 & 1 & 0.5 & 4.5 \\
\midrule
\textbf{Avg} & \textbf{0.50} & \textbf{0.88} & \textbf{0.60} & \textbf{0.83} & \textbf{0.75} & \textbf{0.65} & \textbf{3.70} \\
\textbf{Progress Score} & \multicolumn{7}{c}{\textbf{74.0\%} (= 3.70/5 $\times$ 100\%)} \\
\textbf{Success Rate} & \multicolumn{7}{c}{\textbf{50.0\%} (= 10/20 $\times$ 100\%)} \\
\bottomrule
\end{tabular}
\end{table}

\begin{table}[t]
\centering
\caption{Detailed evaluation results for Fold Clothes task (6 steps, max score 6).}
\label{tab:realworld_detailed_fold_clothes}
\scriptsize
\begin{tabular}{l|c|cccccc|c}
\toprule
\textbf{Trial} & \textbf{Success} & \textbf{Fold Half} & \textbf{Left Sleeve} & \textbf{Right Sleeve} & \textbf{Fold Again} & \textbf{Flatten} & \textbf{Place} & \textbf{Progress} \\
\midrule
\multicolumn{9}{c}{\textbf{Ours}} \\
\midrule
1  & 1 & 1 & 1 & 1 & 1 & 1 & 1 & 6 \\
2  & 0 & 1 & 1 & 0 & 0 & 0 & 0 & 2 \\
3  & 1 & 1 & 1 & 1 & 1 & 0.5 & 0.5 & 5 \\
4  & 1 & 1 & 1 & 1 & 1 & 1 & 1 & 6 \\
5  & 0 & 0.5 & 0 & 0 & 0 & 0 & 0 & 0.5 \\
6  & 0 & 0 & 0 & 0 & 0 & 0 & 0 & 0 \\
7  & 0 & 0.5 & 0 & 0 & 0 & 0 & 0 & 0.5 \\
8  & 1 & 1 & 1 & 1 & 1 & 0.5 & 0.5 & 5 \\
9  & 1 & 1 & 1 & 1 & 1 & 1 & 1 & 6 \\
10 & 1 & 1 & 1 & 1 & 1 & 1 & 1 & 6 \\
11 & 0 & 0.5 & 0 & 0 & 0 & 0 & 0 & 0.5 \\
12 & 0 & 0.5 & 0 & 0 & 0 & 0 & 0 & 0.5 \\
13 & 0 & 0.5 & 0 & 0 & 0 & 0 & 0 & 0.5 \\
14 & 0 & 1 & 1 & 1 & 1 & 1 & 0 & 5 \\
15 & 0 & 0.5 & 0 & 0 & 0 & 0 & 0 & 0.5 \\
16 & 0 & 1 & 1 & 1 & 1 & 0.5 & 0 & 4.5 \\
17 & 0 & 0 & 0 & 0 & 0 & 0 & 0 & 0 \\
18 & 0 & 1 & 1 & 1 & 0.5 & 0 & 0 & 3.5 \\
19 & 0 & 0.5 & 0 & 0 & 0 & 0 & 0 & 0.5 \\
20 & 1 & 1 & 1 & 1 & 1 & 1 & 1 & 6 \\
\midrule
\textbf{Avg} & \textbf{0.35} & \textbf{0.73} & \textbf{0.55} & \textbf{0.50} & \textbf{0.48} & \textbf{0.38} & \textbf{0.30} & \textbf{2.93} \\
\textbf{Progress Score} & \multicolumn{8}{c}{\textbf{48.8\%} (= 2.93/6 $\times$ 100\%)} \\
\textbf{Success Rate} & \multicolumn{8}{c}{\textbf{35.0\%} (= 7/20 $\times$ 100\%)} \\
\midrule
\multicolumn{9}{c}{$\boldsymbol{\pi_{0.5}}$} \\
\midrule
1  & 1 & 1 & 1 & 1 & 1 & 1 & 1 & 6 \\
2  & 1 & 1 & 1 & 1 & 1 & 1 & 1 & 6 \\
3  & 1 & 1 & 1 & 1 & 0.5 & 1 & 0.5 & 5 \\
4  & 0 & 1 & 1 & 1 & 0.5 & 1 & 0 & 4.5 \\
5  & 0 & 0.5 & 1 & 1 & 0.5 & 1 & 0 & 4 \\
6  & 0 & 1 & 1 & 0.5 & 1 & 1 & 0 & 4.5 \\
7  & 0 & 0.5 & 0 & 0 & 0 & 0 & 0 & 0.5 \\
8  & 0 & 0.5 & 0 & 0 & 0 & 0 & 0 & 0.5 \\
9  & 0 & 1 & 1 & 1 & 1 & 1 & 0 & 5 \\
10 & 0 & 0.5 & 1 & 1 & 0.5 & 0 & 0 & 3 \\
11 & 0 & 0.5 & 0 & 0 & 0 & 0 & 0 & 0.5 \\
12 & 0 & 1 & 1 & 1 & 1 & 1 & 0 & 5 \\
13 & 0 & 1 & 1 & 0.5 & 0 & 0 & 0 & 2.5 \\
14 & 1 & 1 & 1 & 1 & 1 & 1 & 1 & 6 \\
15 & 1 & 1 & 1 & 1 & 0.5 & 1 & 1 & 5.5 \\
16 & 0 & 1 & 1 & 1 & 0.5 & 0 & 0 & 3.5 \\
17 & 1 & 1 & 1 & 1 & 1 & 1 & 1 & 6 \\
18 & 0 & 1 & 1 & 1 & 0.5 & 1 & 0 & 4.5 \\
19 & 0 & 0 & 0 & 0 & 0 & 0 & 0 & 0 \\
20 & 0 & 1 & 1 & 1 & 0 & 0 & 0 & 3 \\
\midrule
\textbf{Avg} & \textbf{0.30} & \textbf{0.83} & \textbf{0.80} & \textbf{0.75} & \textbf{0.53} & \textbf{0.60} & \textbf{0.28} & \textbf{3.78} \\
\textbf{Progress Score} & \multicolumn{8}{c}{\textbf{62.9\%} (= 3.78/6 $\times$ 100\%)} \\
\textbf{Success Rate} & \multicolumn{8}{c}{\textbf{30.0\%} (= 6/20 $\times$ 100\%)} \\
\bottomrule
\end{tabular}
\end{table}

\begin{table}[t]
\centering
\caption{Detailed evaluation results for Unpack Delivery task (5 steps, max score 5).}
\label{tab:realworld_detailed_unpack_delivery}
\scriptsize
\begin{tabular}{l|c|ccccc|c}
\toprule
\textbf{Trial} & \textbf{Success} & \textbf{Grab Knife} & \textbf{Push Blade} & \textbf{Handover} & \textbf{Cut Seal} & \textbf{Open Lid} & \textbf{Progress} \\
\midrule
\multicolumn{8}{c}{\textbf{Ours}} \\
\midrule
1  & 1 & 1 & 1 & 1 & 1 & 1 & 5 \\
2  & 1 & 1 & 1 & 1 & 1 & 1 & 5 \\
3  & 1 & 1 & 1 & 1 & 1 & 1 & 5 \\
4  & 0 & 1 & 0 & 0 & 0 & 0 & 1 \\
5  & 0 & 1 & 1 & 1 & 0.5 & 0 & 3.5 \\
6  & 0 & 1 & 1 & 1 & 0 & 0 & 3 \\
7  & 1 & 1 & 1 & 1 & 1 & 1 & 5 \\
8  & 0 & 1 & 1 & 1 & 0 & 0 & 3 \\
9  & 1 & 1 & 1 & 1 & 1 & 1 & 5 \\
10 & 1 & 1 & 1 & 1 & 1 & 1 & 5 \\
11 & 1 & 1 & 1 & 1 & 1 & 1 & 5 \\
12 & 1 & 1 & 1 & 1 & 1 & 1 & 5 \\
13 & 0 & 1 & 1 & 1 & 0 & 0 & 3 \\
14 & 1 & 1 & 1 & 1 & 1 & 1 & 5 \\
15 & 1 & 1 & 1 & 1 & 1 & 1 & 5 \\
16 & 0 & 1 & 1 & 1 & 0 & 0 & 3 \\
17 & 1 & 1 & 1 & 1 & 0.5 & 1 & 4.5 \\
18 & 1 & 1 & 1 & 1 & 1 & 1 & 5 \\
19 & 0 & 1 & 1 & 1 & 0.5 & 0 & 3.5 \\
20 & 1 & 1 & 1 & 1 & 1 & 1 & 5 \\
\midrule
\textbf{Avg} & \textbf{0.65} & \textbf{1.00} & \textbf{0.95} & \textbf{0.95} & \textbf{0.68} & \textbf{0.65} & \textbf{4.23} \\
\textbf{Progress Score} & \multicolumn{7}{c}{\textbf{84.5\%} (= 4.23/5 $\times$ 100\%)} \\
\textbf{Success Rate} & \multicolumn{7}{c}{\textbf{65.0\%} (= 13/20 $\times$ 100\%)} \\
\midrule
\multicolumn{8}{c}{$\boldsymbol{\pi_{0.5}}$} \\
\midrule
1  & 0 & 1 & 1 & 0.5 & 0.5 & 0 & 3 \\
2  & 0 & 1 & 1 & 1 & 0 & 0 & 3 \\
3  & 0 & 1 & 1 & 1 & 0.5 & 0 & 3.5 \\
4  & 0 & 1 & 1 & 1 & 0.5 & 0 & 3.5 \\
5  & 0 & 1 & 1 & 1 & 0.5 & 0 & 3.5 \\
6  & 0 & 1 & 1 & 1 & 0 & 0 & 3 \\
7  & 0 & 1 & 1 & 1 & 0 & 0 & 3 \\
8  & 1 & 1 & 1 & 1 & 0.5 & 1 & 4.5 \\
9  & 0 & 1 & 1 & 1 & 0.5 & 0 & 3.5 \\
10 & 1 & 1 & 1 & 1 & 1 & 1 & 5 \\
11 & 0 & 1 & 1 & 1 & 0.5 & 0 & 3.5 \\
12 & 0 & 1 & 1 & 1 & 0.5 & 0 & 3.5 \\
13 & 1 & 1 & 1 & 1 & 1 & 1 & 5 \\
14 & 0 & 1 & 1 & 1 & 0.5 & 0 & 3.5 \\
15 & 1 & 1 & 1 & 1 & 0.5 & 1 & 4.5 \\
16 & 0 & 1 & 1 & 1 & 0 & 0 & 3 \\
17 & 0 & 1 & 1 & 1 & 0 & 0 & 3 \\
18 & 0 & 1 & 1 & 1 & 0 & 0 & 3 \\
19 & 0 & 1 & 1 & 1 & 0.5 & 0 & 3.5 \\
20 & 1 & 1 & 1 & 1 & 1 & 1 & 5 \\
\midrule
\textbf{Avg} & \textbf{0.25} & \textbf{1.00} & \textbf{1.00} & \textbf{0.98} & \textbf{0.43} & \textbf{0.25} & \textbf{3.65} \\
\textbf{Progress Score} & \multicolumn{7}{c}{\textbf{73.0\%} (= 3.65/5 $\times$ 100\%)} \\
\textbf{Success Rate} & \multicolumn{7}{c}{\textbf{25.0\%} (= 5/20 $\times$ 100\%)} \\
\bottomrule
\end{tabular}
\end{table}

\begin{table}[t]
\centering
\caption{Detailed evaluation results for Insert Tubes task (2 categories: Grasp and Insert, max score 6).}
\label{tab:realworld_detailed_insert_tubes}
\scriptsize
\begin{tabular}{l|c|cc|c}
\toprule
\textbf{Trial} & \textbf{Success} & \textbf{Grasp (3)} & \textbf{Insert (3)} & \textbf{Progress} \\
\midrule
\multicolumn{5}{c}{\textbf{Ours}} \\
\midrule
1  & 0 & 3 & 2 & 5 \\
2  & 1 & 3 & 3 & 6 \\
3  & 0 & 3 & 2 & 5 \\
4  & 0 & 2 & 2 & 4 \\
5  & 1 & 3 & 3 & 6 \\
6  & 1 & 3 & 3 & 6 \\
7  & 0 & 3 & 2 & 5 \\
8  & 1 & 3 & 3 & 6 \\
9  & 0 & 3 & 2 & 5 \\
10 & 0 & 3 & 2 & 5 \\
11 & 1 & 3 & 3 & 6 \\
12 & 0 & 3 & 2 & 5 \\
13 & 1 & 3 & 3 & 6 \\
14 & 0 & 3 & 2 & 5 \\
15 & 0 & 3 & 1 & 4 \\
16 & 1 & 3 & 3 & 6 \\
17 & 0 & 2 & 2 & 4 \\
18 & 0 & 2 & 2 & 4 \\
19 & 1 & 3 & 3 & 6 \\
20 & 0 & 2 & 2 & 4 \\
\midrule
\textbf{Avg} & \textbf{0.40} & \textbf{2.80} & \textbf{2.35} & \textbf{5.15} \\
\textbf{Progress Score} & \multicolumn{4}{c}{\textbf{85.8\%} (= 5.15/6 $\times$ 100\%)} \\
\textbf{Success Rate} & \multicolumn{4}{c}{\textbf{40.0\%} (= 8/20 $\times$ 100\%)} \\
\midrule
\multicolumn{5}{c}{$\boldsymbol{\pi_{0.5}}$} \\
\midrule
1  & 0 & 3 & 1 & 4 \\
2  & 0 & 3 & 1 & 4 \\
3  & 0 & 3 & 1 & 4 \\
4  & 0 & 3 & 1 & 4 \\
5  & 0 & 3 & 1 & 4 \\
6  & 0 & 3 & 1 & 4 \\
7  & 0 & 3 & 1 & 4 \\
8  & 0 & 3 & 2 & 5 \\
9  & 1 & 3 & 3 & 6 \\
10 & 1 & 3 & 3 & 6 \\
11 & 1 & 3 & 3 & 6 \\
12 & 0 & 2 & 1 & 3 \\
13 & 0 & 3 & 2 & 5 \\
14 & 0 & 2 & 2 & 4 \\
15 & 1 & 3 & 3 & 6 \\
16 & 1 & 3 & 3 & 6 \\
17 & 0 & 3 & 2 & 5 \\
18 & 0 & 2 & 2 & 4 \\
19 & 1 & 3 & 3 & 6 \\
20 & 0 & 3 & 2 & 5 \\
\midrule
\textbf{Avg} & \textbf{0.30} & \textbf{2.85} & \textbf{1.90} & \textbf{4.75} \\
\textbf{Progress Score} & \multicolumn{4}{c}{\textbf{79.2\%} (= 4.75/6 $\times$ 100\%)} \\
\textbf{Success Rate} & \multicolumn{4}{c}{\textbf{30.0\%} (= 6/20 $\times$ 100\%)} \\
\bottomrule
\end{tabular}
\end{table}

\begin{table}[t]
\centering
\caption{Detailed evaluation results for Fold Pants task (3 steps, max score 3).}
\label{tab:realworld_detailed_fold_pants}
\scriptsize
\begin{tabular}{l|c|ccc|c}
\toprule
\textbf{Trial} & \textbf{Success} & \textbf{Fold 1} & \textbf{Fold 2} & \textbf{Place} & \textbf{Progress} \\
\midrule
\multicolumn{6}{c}{\textbf{Ours}} \\
\midrule
1  & 1 & 1 & 1 & 1 & 3 \\
2  & 1 & 1 & 1 & 1 & 3 \\
3  & 1 & 1 & 1 & 1 & 3 \\
4  & 0 & 0 & 0 & 0 & 0 \\
5  & 0 & 1 & 0 & 0 & 1 \\
6  & 0 & 1 & 0 & 0 & 1 \\
7  & 1 & 1 & 1 & 1 & 3 \\
8  & 1 & 1 & 1 & 1 & 3 \\
9  & 1 & 1 & 1 & 1 & 3 \\
10 & 1 & 1 & 1 & 1 & 3 \\
11 & 1 & 1 & 1 & 1 & 3 \\
12 & 1 & 1 & 1 & 1 & 3 \\
13 & 1 & 1 & 1 & 1 & 3 \\
14 & 1 & 1 & 1 & 1 & 3 \\
15 & 1 & 1 & 1 & 1 & 3 \\
16 & 1 & 1 & 1 & 1 & 3 \\
17 & 1 & 1 & 1 & 1 & 3 \\
18 & 0 & 1 & 0 & 0 & 1 \\
19 & 0 & 1 & 0 & 0 & 1 \\
20 & 0 & 0 & 0 & 0 & 0 \\
\midrule
\textbf{Avg} & \textbf{0.70} & \textbf{0.90} & \textbf{0.70} & \textbf{0.70} & \textbf{2.30} \\
\textbf{Progress Score} & \multicolumn{5}{c}{\textbf{76.7\%} (= 2.30/3 $\times$ 100\%)} \\
\textbf{Success Rate} & \multicolumn{5}{c}{\textbf{70.0\%} (= 14/20 $\times$ 100\%)} \\
\midrule
\multicolumn{6}{c}{$\boldsymbol{\pi_{0.5}}$} \\
\midrule
1  & 1 & 1 & 1 & 1 & 3 \\
2  & 1 & 1 & 1 & 1 & 3 \\
3  & 1 & 1 & 1 & 1 & 3 \\
4  & 0 & 0 & 0 & 0 & 0 \\
5  & 0 & 0 & 0 & 0 & 0 \\
6  & 0 & 0 & 0 & 0 & 0 \\
7  & 0 & 0 & 0 & 0 & 0 \\
8  & 0 & 0 & 0 & 0 & 0 \\
9  & 0 & 0 & 0 & 0 & 0 \\
10 & 0 & 0 & 0 & 0 & 0 \\
11 & 1 & 1 & 1 & 1 & 3 \\
12 & 1 & 1 & 1 & 1 & 3 \\
13 & 1 & 1 & 1 & 1 & 3 \\
14 & 0 & 0 & 0 & 0 & 0 \\
15 & 0 & 0 & 0 & 0 & 0 \\
16 & 0 & 0 & 0 & 0 & 0 \\
17 & 0 & 0 & 0 & 0 & 0 \\
18 & 0 & 0 & 0 & 0 & 0 \\
19 & 0 & 0 & 0 & 0 & 0 \\
20 & 0 & 0 & 0 & 0 & 0 \\
\midrule
\textbf{Avg} & \textbf{0.30} & \textbf{0.30} & \textbf{0.30} & \textbf{0.30} & \textbf{0.90} \\
\textbf{Progress Score} & \multicolumn{5}{c}{\textbf{30.0\%} (= 0.90/3 $\times$ 100\%)} \\
\textbf{Success Rate} & \multicolumn{5}{c}{\textbf{30.0\%} (= 6/20 $\times$ 100\%)} \\
\bottomrule
\end{tabular}
\end{table}

\end{document}